\documentclass[journal]{IEEEtran}
\ifCLASSINFOpdf
\else
\fi
\long\def\comment#1{}

\newfont{\bbb}{msbm10 scaled 700}

\newfont{\bb}{msbm10 scaled 1100}

\newcommand{\RR}{\mbox{\bb R}}

\newcommand{\EE}{\mbox{\bb E}}

\newcommand{\pv}{{\bf p}}

\newcommand{\uv}{{\bf u}}

\newcommand{\vv}{{\bf v}}
\newcommand{\xv}{{\bf x}}

\newcommand{\zv}{{\bf z}}
\newcommand{\zerov}{{\bf 0}}

\newcommand{\Id}{{\bf I}}

\newcommand{\Ac}{{\cal A}}

\newcommand{\Fc}{{\cal F}}

\newcommand{\Hc}{{\cal H}}

\newcommand{\Lc}{{\cal L}}

\newcommand{\Nc}{{\cal N}}

\newcommand{\Tc}{{\cal T}}

\newcommand{\Vc}{{\cal V}}
\newcommand{\Xc}{{\cal X}}
\newcommand{\Yc}{{\cal Y}}

\newcommand{\alphav}{\hbox{\boldmath$\alpha$}}

\newcommand{\thetav}{\hbox{\boldmath$\theta$}}

\renewcommand{\arg}{{\hbox{arg}}}

\newcommand{\eqdef}{\stackrel{\Delta}{=}}

\newcommand{\trasp}{{\sf T}}

\usepackage{array}
\usepackage{multirow}
\usepackage{enumerate}
\usepackage{makecell}
\usepackage{tabularx} 
\usepackage{algorithm,algcompatible}
\usepackage{lipsum}
\usepackage{bbm}
\usepackage{times}
\usepackage[cmex10]{amsmath}
\usepackage{algpseudocode}
\usepackage{epsfig,verbatim}
\usepackage{amssymb} 
\usepackage{graphics,subfigure}
\usepackage{xcolor}
\usepackage{graphicx}
\usepackage{tabularx,booktabs}

\newtheorem{theorem}{Theorem}

\newtheorem{lemma}{Lemma}

\newtheorem{remark}{Remark}

\newtheorem{example}{Example}

\begin{document}
%
\title{Active Learning with Multiple Kernels}
%
%
%

\author{Songnam Hong,~\IEEEmembership{Member,~IEEE,}
        and~Jeongmin Chae,~\IEEEmembership{Student,~IEEE}
\thanks{S. Hong and J. Chae are with the Department of Electrical and Computer Engineering, Ajou University, Suwon, 16499, Korea. (e-mail:\{snhong, jmchae92\}@ajou.ac.kr)}.
}

%
%




\maketitle

\begin{abstract}
Online multiple kernel learning (OMKL) has provided an attractive performance in nonlinear function learning tasks. Leveraging a random feature approximation, the major drawback of OMKL, known as the curse of dimensionality, has been recently alleviated. In this paper, we introduce a new research problem, termed (stream-based) {\em active multiple kernel learning} (AMKL), in which a learner is allowed to label selected data from an oracle according to a selection criterion. This is necessary in many real-world applications as acquiring true labels is costly or time-consuming. We prove that AMKL achieves an optimal sublinear regret $\mathcal{O}(\sqrt{T})$, implying that the proposed selection criterion indeed avoids unuseful label-requests. Furthermore, we propose AMKL with an adaptive kernel selection (AMKL-AKS) in which irrelevant kernels can be excluded from a kernel dictionary `on the fly'.  This approach can improve the efficiency of active learning as well as the accuracy of a function approximation. Via numerical tests with various real datasets, it is demonstrated that AMKL-AKS yields a similar or better performance than the best-known OMKL, with a smaller number of labeled data.

\end{abstract}


%
\IEEEpeerreviewmaketitle

%
%
\section{Introduction}\label{sec:intro}


Learning a non-linear function is of great interest in various machine learning tasks as classification, regression, clustering, dimensionality reduction, and reinforcement learning \cite{scholkopf2001learning,shawe2004kernel,lin2010multiple, dai2016learning}. In particular, supervised functional learning tasks, which are closely related to the subject of this paper, are  formulated as follows. Given data samples $\{(\xv_t,y_t): t=1,...,T\}$ with features $\xv_t \in \RR^d$ and labels $y_t \in \RR$, the goal is to learn a function $f:\RR^d\rightarrow \RR$ such that each pair of true label $y_t$ and estimated label $\hat{f}(\xv_t)$ is minimized. This challenge problem can be tractable with the restriction that $f(\cdot)$ belongs to a reproducing kernel Hilbert space (RKHS) \cite{scholkopf2001learning}. The performance of this kernel-based learning completely relies on a preselected kernel, which is determined either manually based on a task-specific priori knowledge or by some intensive cross-validation process. Multiple kernel learning (MKL), using a predefined set of kernels (i.e., a kernel dictionary), is more powerful as it can enable a data-driven kernel selection from a given dictionary \cite{shawe2004kernel, rakotomamonjy2008simplemkl, cortes2012l2, gonen2011multiple, bazerque2013nonparametric}. Specifically, MKL seeks an optimal linear (or non-linear) combination of multiple kernels as part of an learning algorithm.

In many real-world applications, Learning tasks are  expected to be performed in an {\em online} fashion. For example, online learning is required when data arrive sequentially such as online spam detection \cite{ma2009identifying} and time series prediction \cite{richard2008online}, and when the large number of data makes it impossible to carry out data analytic in batch form \cite{kivinen2004online}. For such cases, 
online MKL (OMKL) has been proposed, which seeks the optimal combination of a pools of single kernel functions in an online fashion. It can  yield a superior accuracy and enjoy a great flexibility compared with single-kernel online learning \cite{kivinen2004online,sahoo2014online,shen2019random}. In contrast, OMKL generally suffers from a high computational complexity since the dimension of optimization variables grow with time (i.e., the number of data samples) \cite{wahba1990spline, shawe2004kernel}. Recently in \cite{shen2019random}, this problem, known as the curse of dimensionality, has been alleviated by applying random feature (RF) approximation \cite{rahimi2008random} to OMKL framework. In RF-based OMKL, the dimension of optimization can be controlled by taking into account the accuracy-complexity tradeoff. Another advantage of RF-based OMKL is that a function approximation can be solved using the powerful toolboxes from online convex optimization and online learning in vector spaces \cite{shen2019random}.

\begin{figure}[!t]
\centerline{\includegraphics[width=8.5cm]{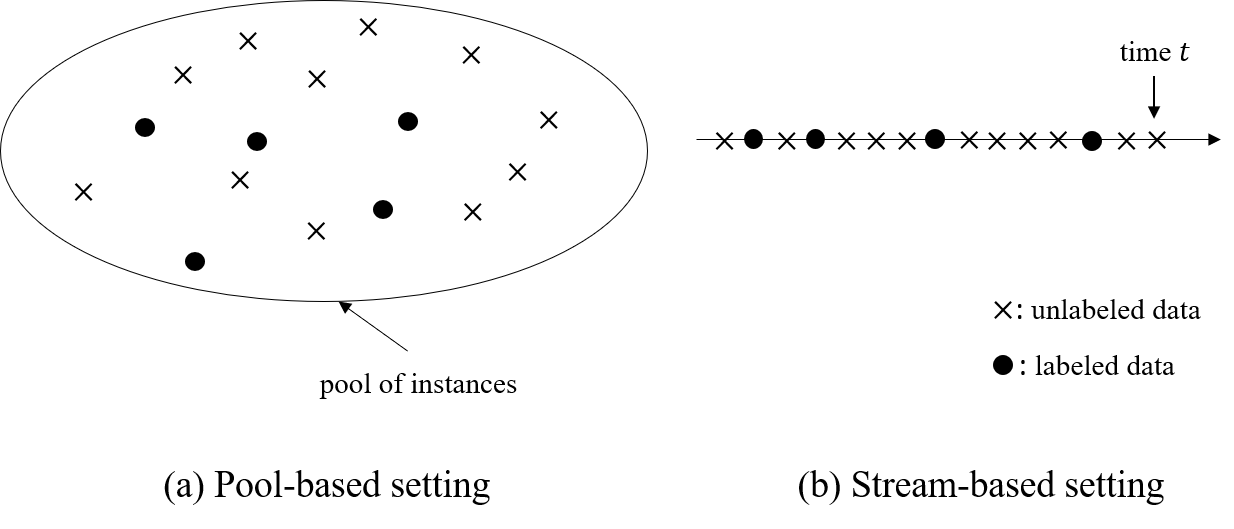}}
\caption{Two types of active learning. At time $t$, the learner can query the oracle for the label of any unlabeled data in the pool-based setting, whereas in the stream-based setting, the learner can query the oracle only for the label of an incoming unlabeled data.}
\label{fig:AL}
\end{figure}

Unlabeled data may be abundant but labels are difficult, time-consuming, or expensive to acquire, especially when only experts whose time is precious can provide reliable labels \cite{zhu2005semi, settles2008active, bordes2005fast}. Active learning aims at overcoming the labeling bottleneck by allowing the learner to decide whether or not to acquire the label of an incoming data from the oracle. In this way, the learner can achieve high accuracy by using as few labeled data as possible, thereby minimizing the cost of obtaining labeled data. There are various scenarios in which the learner may be able to ask queries. Based on different ways of entering the unlabeled data, active learning can be categorized into pool-based \cite{sugiyama2009pool, mccallumzy1998employing} and stream-based models \cite{smailovic2014stream, settles2009active, wu2018pool}, as illustrated in Fig.~\ref{fig:AL}.
In particular for OMKL framework to be focused in this paper, {\em stream-based (or sequential) active learning} is most relevant since each unlabeled data is typically drawn one at a time from the data source, and the learner must decide whether to query or discard it \cite{settles2009active}.  The stream-based scenario has been studied in several machine learning tasks such as speech tagging \cite{dagan1995committee}, sensor scheduling \cite{krishnamurthy2002algorithms}, information retrieval \cite{yu2005svm}, drifting streaming data \cite{vzliobaite2013active} and expert advice \cite{hao2018online}. Despite its practical necessity, active learning has not been investigated under OMKL frameworks.

{\bf Our contributions:} We propose a stream-based active learning for OMKL frameworks, which is referred to as {\em active multiple kernel learning} (AMKL). In the proposed AMKL, a learner is allowed to label selected incoming data according to a selection criterion, whereas in OMKL, all incoming data are assumed to be labeled. The proposed selection criterion guarantees that unlabeling of an incoming data only causes a $\eta_c$-bounded loss compared with OMKL counterpart. Here, the parameter $\eta_c$ can be chosen by considering the tradeoff of active-learning efficiency and function-approximation accuracy. We prove that AMKL with $\eta_c = \mathcal{O}(1/\sqrt{T})$ achieves an optimal sublinear regret $\mathcal{O}(\sqrt{T})$, implying that the proposed selection criterion indeed avoids unnecessary label-requests. In multiple kernel learning as AMKL and OMKL, the use of a large kernel dictionary may deteriorate  the accuracy of a function learning or cause a slower convergence to an optimal function if too many irrelevant kernels are included. We address this problem by presenting an adaptive kernel selection for AMKL and OMKL (termed AMKL-AKS and OMKL-AKS). The proposed kernel selection can rule out irrelevant kernels `on the fly', where they are determined on the basis of accumulated loss function. Also, it is a randomized algorithm which can ensure a sort of exploration and provide the robustness to potential adversarial attacks. Using  a martingale argument, we prove that both AMKL-AKS and OMKL-AKS keeps the optimal sublinear regret with high probability. More importantly, the proposed kernel selection can improve active-learning efficiency considerably, by enabling that the selection criterion is checked only with refined relevant kernels. Without this, the irrelevant kernels generate inaccurate outputs, which hinders satisfying the selection criterion irrespective of the usefulness of labeling. Via numerical tests with real datasets, it is verified that AMKL-AKS provides a similar or better accuracy than the best-known OMKL (a.k.a., Raker), with a smaller number of labeled samples. Thus, the proposed method can yield an elegant accuracy-efficiency tradeoff.

The remainder of this paper is organized as follows. In Section~\ref{sec:pre}, we briefly review RF-based MKL, which is the underlying method of the proposed algorithms. In Section~\ref{sec:Alg}, we describe the proposed active learning algorithms, named AMKL and AMKL-AKS. Regret analysis is provided in Section~\ref{sec:regret} to show the asymptotic optimality of the proposed methods. In Section~\ref{sec:Exp}, beyond the asymptotic analysis, their superiority are verified via numerical tests with various real datasets. Some concluding remarks are provided in Section~\ref{sec:con}.


{\em Notations:} Bold lowercase letters  will denote column vectors. For any vector $\xv$, $\xv^{\trasp}$ stands for the transpose of $\xv$ and $\|\xv\|$ denotes the $\ell_2$-norm of $\xv$. $\EE[\cdot]$ denotes the expectation and $\langle\cdot,\cdot\rangle$ denotes the inner product in Euclidean space. To simplify notations, we let $[N]\eqdef\{1,2,...,N\}$ for any positive integer $N$.

%
%

\section{Preliminaries}\label{sec:pre}
We briefly review multiple kernel learning (MKL) based on random feature (RF) approximation as it is the baseline method for the proposed online MKL (OMKL) and active (AMKL) algorithms. Given the training samples $\{(\xv_1,y_1),...,(\xv_{T},y_T)\}$, where $\xv_{t}\in \Xc \subseteq \RR^d$ and $y_t \in \Yc\subseteq \RR$, the goal is to learn a non-linear function $f: \Xc \rightarrow \Yc$ such that $f(\xv_{t})=y_t$ for $t\in [T]$. In kernel-based learning \cite{smola1998learning, gonen2011multiple, bazerque2013nonparametric}, it is assumed that a target function $f(\xv)$ belongs to a reproducing Hilbert kernel space (RKHS), defined as
\begin{equation}
    \Hc \eqdef \{f: f(\xv)=\sum_{t=1}^{\infty} \alpha_t \kappa(\xv,\xv_t)\},
\end{equation} where $\kappa(\xv,\xv_t): \Xc \times \Xc \rightarrow \Yc$ is a symmetric positive semidefinite basis function (called kernel), which measures the similarity between $\xv$ and $\xv_t$. Among various kernels, one representative example is the Gaussian kernel with a parameter $\sigma^2$, given as
\begin{equation} \label{eq:Gaussian_kernel}
    \kappa(\xv,\xv_t)\eqdef \exp\left(\frac{-\|\xv-\xv_t\|^2}{2\sigma^2}\right).
\end{equation} Also, a kernel is said to be reproducing if the following holds:
\begin{equation}
    \langle\kappa(\xv,\xv_{t}),\kappa(\xv,\xv_{t'})\rangle_{\Hc} = \kappa(\xv_t,\xv_{t'}),
\end{equation} where $\langle\cdot,\cdot\rangle_{\Hc}$ denotes an inner product defined in the Hilbert space $\Hc$. Also, the associated RKHS norm is defined as 
\begin{equation}\label{eq:kernel_fn}
    \|f\|_{\Hc}^2 \eqdef \sum_{t}\sum_{t'}\alpha_t\alpha_{t'}\kappa(\xv_{t},\xv_{t'}).
\end{equation} The function approximation problem over RHKS is formulated as
\begin{equation}\label{eq:opt}
    \min_{f\in\Hc}\frac{1}{T} \sum_{t=1}^{T}\Lc(f(\xv_t),y_t),
\end{equation}where $\Lc(\cdot,\cdot)$ stands for a loss function. Note that this loss function can be task-specific, e.g.,  least-square for regression and logistic cost for classification.

Especially when the number of data samples is finite (e.g., $T$ training samples), the representer theorem in \cite{wahba1990spline} shows that the optimal solution of (\ref{eq:opt}) is represented as
\begin{equation}\label{eq:appr}
    \hat{f}(\xv) = \sum_{t=1}^{T}\alpha_t \kappa(\xv,\xv_t).
\end{equation} The major drawback of this approach is the curse of dimensionality as the number of parameters $\alpha_t$'s (to be optimized) grows with the number of data samples $T$.

In \cite{rahimi2008random}, it has been addressed by introducing RF approximation for kernels. As in \cite{rahimi2008random}, the kernel $\kappa$ in (\ref{eq:kernel_fn}) is assumed to be shift-invariant, that is,  $\kappa(\xv_t,\xv_{t'})=\kappa(\xv_t - \xv_t')$. Note that Gaussian, Laplacian, and Cauchy kernels satisfy the shift-invariant \cite{rahimi2008random}. For $\kappa(\xv_t - \xv_t')$ absolutely integrable, its Fourier transform $\pi_{k}(\vv)$ exists and represents the power spectral density. Also, when $\kappa(\zerov)=1$ it can also be viewed as a probability density function (PDF). For a Gaussian kernel in (\ref{eq:Gaussian_kernel}), we have $\pi_{\kappa}(\vv)=\Nc(0,\sigma^{-2}\Id)$. Then, we have:
\begin{equation}
    \kappa(\xv_{t}-\xv_{t'}) = \EE\left[\exp\left(j\vv^T(\xv_t-\xv_{t'})\right)\right].
\end{equation} Having a sufficient number of independent and identically distributed (i.i.d.) samples $\{\vv_i:i\in[D]\}$ from $\pi_{\kappa}(\vv)$, $\kappa(\xv_{t}-\xv_{t'})$ can be well-approximated by the sample mean as
\begin{equation}\label{RF_AP}
    \kappa(\xv_{t} - \xv_{t'}) \approx \frac{1}{D}\sum_{i=1}^{D} \mbox{Re}\left(\exp\left(j\vv_{i}^T(\xv_t-\xv_{t'})\right)\right),
\end{equation}where $\mbox{Re}(a)$ denotes the real part of a complex value $a$. Clearly, the accuracy of this approximation grows as the number of samples $D$ increases. In numerical tests, a proper $D$ will be chosen by considering the accuracy-complexity tradeoff. The approximation in (\ref{RF_AP}) can be rewritten as vector form:
\begin{equation}
    \kappa(\xv_{t} - \xv_{t'}) = \zv^{\trasp}(\xv_t)\zv(\xv_{t'}),
\end{equation} where
\begin{equation}\label{eq:zv}
    \zv(\xv)=\frac{1}{\sqrt{D}}\left[\sin{\vv_1^{\trasp}\xv},...,\sin{\vv_{D}^{\trasp}(\xv),\cos{\vv_{1}^{\trasp}\xv},...,\cos{\vv_{D}^{\trasp}\xv}}\right]^{\trasp}.
\end{equation} Based on this, the optimal solution $\hat{f}(\xv)$ in (\ref{eq:appr}) can be well-approximated as
\begin{equation}\label{eq:RFA}
    \hat{f}(\xv) = \sum_{t=1}^{T}\alpha_t\zv^{\trasp}(\xv_t)\zv(\xv)  \eqdef \hat{\thetav}^{\trasp}\zv(\xv),
\end{equation} where the optimization variable $\hat{\thetav}$ is a $2D$-vector. Note that its dimension $2D$ can be determined irrespective of the number of data samples (e.g., $T$).

RF-based kernel learning can be naturally extended into MKL framework, where a target function is formed as a linear (or convex) combination of multiple preselected kernels $\{\kappa_i: i\in[P]\}$. From \cite{micchelli2005learning}, the function approximation can be represented as
\begin{equation}\label{eq:MKL_f}
    \hat{f}(\xv)=\sum_{i=1}^{P} \hat{p}_i \hat{f}_i (\xv) \in \bar{\Hc},
\end{equation} where $\bar{\Hc}\eqdef \Hc_1 \bigotimes \Hc_2 \bigotimes \cdots \bigotimes \Hc_P$ and
$\hat{f}_i(\xv) \in \Hc_{i}$ which is a RKHS induced by the kernel $\kappa_i$, and $\hat{p}_i\in[0,1]$ denotes the combination weight of the associated kernel function $\hat{f}_i$. Also, under RF approximation, the kernel functions in (\ref{eq:MKL_f}) can be further simplified as
\begin{equation}\label{eq:kernel_form}
    \hat{f}_{i}(\xv) = \hat{\thetav}_i^{\trasp} \zv_{i}(\xv),
\end{equation}for $i\in[P]$, where $\zv_{i}$ is defined as in (\ref{eq:zv}) with $D$ number of i.i.d. samples from $\pi_{\kappa_i}(\vv)$. 
Assuming RF-based MKL, kernel functions in the following sections have the forms of
 (\ref{eq:kernel_form}).

%
%
\section{Methods}\label{sec:Alg}

We first formulate the problem setting for online MKL (OMKL). The main purpose is to learn a sequence of functions $\hat{f}_{t+1}(\xv)$, $t\in [T]$, in an {\em online} fashion. Specifically, at each time $t$, a function $\hat{f}_{t+1}: \Xc \times \Yc$ is learned from the training samples $\{(\xv_{\tau},y_{\tau}): \tau\in[t]\}$, where the feature vector $\xv_{\tau} \in \Xc \subseteq \RR^d$ and the label $y_{\tau} \in \Yc\subseteq \RR$. We let $\Lc: \Xc\times \Xc \rightarrow \RR$ denote a loss function to evaluate the accuracy of a learned function. Throughout the paper, it is assumed that there are $P$ kernels in a kernel dictionary. OMKL frameworks consist of two steps, called {\em local} and {\em global} steps. In local step, each kernel function $\hat{f}_{t+1,i}(\xv)$ is optimized independently from the other kernel functions. In global step, the learner seeks the best function approximation 
$\hat{f}_{t+1}(\xv)$ by combining the kernel functions $\{\hat{f}_{t+1,i}(\xv), i\in [P]\}$ with some weights $\{\hat{p}_{t+1}(i), i\in[P]\}$, i.e.,
\begin{equation}
    \hat{f}_{t+1}(\xv) = \sum_{i=1}^{P} \hat{p}_{t+1}(i)\hat{f}_{t+1,i}(\xv).
\end{equation}  Our goal is to learn a sequence of functions $\hat{f}_{t}(\xv)$ that minimize (cumulative) {\em regret}, defined as
\begin{equation}
    {\rm regret}_{T} = \sum_{t=1}^{T} \Lc(\hat{f}_{t}(\xv_t),y_t) - \min_{1\leq i\leq P} \sum_{t=1}^{T} f^{\star}_{i}(\xv_t),
\end{equation} where the regret compares the cumulative loss of the learner to the cumulative loss of the best kernel in hindsight.

In Section~\ref{subset:OMKL}, we improve (RF-based) OMKL frameworks by refining irrelevant kernels efficiently. Then, in Section~\ref{subsec:AMKL}, we propose novel stream-based active learning for OMKL frameworks, in which a learner is allowed to label selected incoming data from an oracle.

\subsection{The Proposed OMKL-AKS}\label{subset:OMKL}

We propose OMKL with an adaptive kernel selection (OMKL-AKS), in which the overall $P$ kernels are refined at every time $t$ on the basis of the accumulated loss information. This approach is motivated by the fact that using a large kernel dictionary can deteriorate the accuracy of a function approximation if too many irrelevant kernels are included. By excluding such kernels efficiently, OMKL-AKS can provide a faster convergence to an optimal function approximation. The proposed kernel selection is a {\em randomized} algorithm rather than selecting a fixed number of kernels having the lowest accumulated losses. The merits of the randomized approach is following: i) it can ensure a sort of exploration to overcome the low-reliability of the loss values at the beginning of learning process; ii) it can provide a more robustness to a potential adversarial attack.

Focusing on time $t$, the (RF-based) OMKL is first explained and based on this, the proposed OMKL-AKS is described. Note that during the previous time slots, the learner is aware of the  kernel functions $\hat{f}_{t,i}(\xv)$ (i.e., the parameter $\hat{\thetav}_{t,i}$) and the losses $\{\Lc(\hat{f}_{\tau,i}(\xv_{\tau}), y_{\tau}): \tau\in[t]\}$. The OMKL framework consists of the following two steps.

\vspace{0.1cm}
{\em i) Local step:} This step learns a set of single kernel functions $\hat{f}_{t+1,i}(\xv) \in \Hc_i$ for $i \in [P]$. To be specific, the learner observes the new training sample $(\xv_{t},y_{t})$ and optimizes the best kernel approximation $\hat{f}_{t+1,i}(\xv)$ for $i\in [P]$, via an online optimization.
Following the RF approximation in (\ref{eq:RFA}), each kernel function is fully determined by $2D$-vector $\hat{\thetav}_{t+1,i}$ as
\begin{equation}\label{eq:single_k}
    \hat{f}_{t+1,i}(\xv) = \hat{\thetav}_{t+1,i}^{\trasp}\zv_{i}(\xv), 
\end{equation}for $i\in[P]$, where $\zv_{i}(\xv)$ is defined in (\ref{eq:zv}). In this paper, the parameter vector $\hat{\thetav}_{t+1,i}$ is optimized via
 the well-known {\em online gradient descent} (OGD) \cite{hazan2016introduction} as
\begin{equation}\label{eq:OCO}
    \hat{\thetav}_{t+1,i} = \hat{\thetav}_{t,i} - \eta_{l}\nabla\Lc\left(\hat{\thetav}_{t,i}^{\trasp}\zv_{i}(\xv_{t}),y_{t}\right),
\end{equation} for $i\in[P]$, where $\nabla\Lc(\hat{\thetav}_{t,i}^{\trasp}\zv_{i}(\xv_{t}),y_{t})$ denotes the gradient at $\hat{\thetav}=\hat{\thetav}_{t,i}$. As an example, the regularized least-square loss function is defined as 
\begin{equation}
    \Lc(\thetav_{t,i}^{\trasp}\zv_{i}(\xv_{t}), y_{t})= \left[y_{t} - \thetav_{t,i}^{\trasp}\zv_{i}(\xv_{t}))\right]^2 + \lambda \|\thetav_{t,i}\|^2,
\end{equation}with a regularization parameter $\lambda>0$. Then, the gradient is computed as
\begin{equation*}
    \nabla\Lc\left(\hat{\thetav}_{t,i}^{\trasp}\zv_{i}(\xv_{t}),y_{t}\right)= 2\left(\thetav_{t,i}^{\trasp}\zv_{i}(\xv_{t})) - y_t\right)\zv_{i}(\xv_{t}) + 2\lambda\thetav_{t,i}.
\end{equation*}

\vspace{0.1cm}
{\em ii) Global step:} This step learns a target function $\hat{f}_{t+1}(\xv)$ by properly combining the single kernel functions $\{\hat{f}_{t+1,i}(\xv): i \in [P]\}$ such as
\begin{equation}\label{eq:OMKL_f}
    \hat{f}_{t+1}(\xv)= \sum_{i=1}^{P} \hat{p}_{t+1}(i) \hat{f}_{t+1,i} (\xv),
\end{equation} where $\hat{p}_{t+1}(i) \in [0,1]$ represents the combination weight of the kernel function $i$. The learned function can generate the label of an incoming data $\xv_{t+1}$ as
\begin{equation}
    \hat{y}_{t+1}=\hat{f}_{t+1}(\xv_{t+1}).
\end{equation} 
Definitely, the choice of the weight vector $\hat{\pv}_{t+1}=(\hat{p}_{t+1}(1),...,\hat{p}_{t+1}(P))^{\trasp}$ plays a key role in determining the accuracy of OMKL algorithm (i.e., the learned function $\hat{f}_{t+1}(\xv)$). In the context of online learning, the so-called {\em exponential strategy} (EXP strategy) is widely used \cite{bubeck2011introduction}, where the weights are determined on the basis of the past losses as
\begin{equation}\label{eq:exp_st}
    \hat{p}_{t+1}(i) \eqdef \frac{\hat{w}_{t+1}(i)}{\sum_{i=1}^{P}\hat{w}_{t+1}(i)},
\end{equation} for some parameter $\eta_g>0$, where the initial values are $w_{1}(i)=1$ and
\begin{equation} \label{eq:o_w}
\hat{w}_{t+1}(i)=\exp\left(-\eta_{g}\sum_{\tau=1}^{t}\Lc(\hat{f}_{i,\tau}(\xv_\tau),y_\tau)\right),
\end{equation} for $i\in[P]$. 
The OMKL algorithm, based on RF approximation and EXP strategy, is also known as Raker \cite{shen2019random}.

From now on, we describe the proposed OMKL-AKS which can enhance the accuracy of a function approximation by refining kernels adaptively. At every time $t$, the proposed method only employs a subset of $P$ kernels, which is determined on the basis of the weight information $\hat{p}_{t+1}(i), i\in[P]$ (i.e., the accumulated loss information). We first introduce a design parameter $K_{t+1} \in [P]$ which indicates the number of kernels to be used for the construction of a function $\hat{f}_{t+1}(\xv)$. Clearly, the proposed method can include the RF-based OMKL (a.k.a., Raker) as a special case, by setting $K_{t}=P$ for all $t\in[T]$. We suggest a reasonable way to choose the parameter $K_{t+1}$ as
\begin{equation}\label{eq:d_r_t}
    K_{t+1} = \left|\left\{i\in[P]: \hat{p}_{t+1}(i)/\hat{p}^{\star}_{t+1} > \delta_{t+1} \right\}\right|,
\end{equation} for some parameter $\delta_{t+1}$, where $\hat{p}^{\star}_{t+1}=\max_{j\in [P]} \hat{p}_{t+1}(j)$. This approach will be used for our experiments. We would like to remark that OMKL-AKS with any choice of $K_{t+1}$ can guarantee the optimal sublinear regret (i.e., the optimal asymptotic performance).

Given $K_{t+1}$, define the collection of all size-$K_{t+1}$ subsets of $[P]$ as
\begin{align}
\Omega(K_{t+1}) &=\{\Vc: \Vc \subseteq [P], |\Vc|=K_{t+1}\} \nonumber\\
&\eqdef \{\Vc_1,...,\Vc_{|\Omega(K_{t+1})|}\}, \label{eq:collection}
\end{align} where $|\Omega(K_{t+1})|={P \choose K_{t+1}}$. It is noticeable that each kernel index occurs uniformly in the collection  $\Omega(K_{t+1})$.  This property is referred to as {\em uniform frequency}. Also, the corresponding frequency, denoted by $J_{t+1}$, is computed as
\begin{equation}\label{eq:dl}
    J_{t+1}=\frac{K_{t+1} { P \choose K_{t+1}}}{P}, 
\end{equation} since $P\cdot J_{t+1} = |\Omega(K_{t+1})| \cdot K_{t+1}$. By construction, $J_{t+1}$ in (\ref{eq:dl}) should be an integer. In the example of $P=4$ and $K_{t+1}=2$, we have:
\begin{equation*}
    \Omega(K_{t+1}=2)=\{(1,2),(1,3),(1,4),(2,3),(2,4),(3,4)\},
\end{equation*} where each kernel index occurs exactly $J_{t+1}=3$ times, thus satisfying the uniform frequency. Then, a size-$K_{t+1}$ subset is chosen randomly from  $\Omega(K_{t+1})$ according to a certain probability distribution. The specific selection procedure will be explained at the bottom of this section. One may concern the complexity problem to generate all the subsets belong to $\Omega(K_{t+1})$, especially for a large $P$. To address this problem, we choose $J_{t+1}=\gamma_{t+1}K_{t+1}$ with a parameter $\gamma_{t+1}$ such that $J_{t+1}$ is an integer, where $\gamma_{t+1}$ is chosen by considering the size of the collection.   Given $J_{t+1}$ and $K_{t+1}$,  define a collection $\Omega(J_{t+1},K_{t+1})$ whose size is determined as
\begin{equation}\label{eq:col-size}
     |\Omega(J_{t+1},K_{t+1})| \eqdef  \lfloor J_{t+1}\cdot P/ K_{t+1} \rfloor = \lfloor \gamma_{t+1}P \rfloor,
\end{equation} where $\lfloor x \rfloor$ denotes a floor function which produces the greatest integer less than or equal to $x$. Although there might be various methods to construct the elements (i.e., the subsets of $[P]$) of  $\Omega(J_{t+1},K_{t+1})$, the experiments in this paper use a simple balls-bins random construction in Remark~\ref{rem:RC}.


\begin{remark}\label{rem:RC}(Balls-Bins Construction)
Given $J_{t+1}$ and $K_{t+1}$, the elements of $\Omega(J_{t+1},K_{t+1})$ are determined via Balls-Bins construction. Here, kernels and subsets (i.e., elements of $\Omega(J_{t+1},K_{t+1})$) correspond to balls and bins, respectively. Then, there are $P$ balls and $|\Omega(J_{t+1},K_{t+1})|$ bins. As in well-known balls and bins problem, consider the process of tossing $P$ balls into $|\Omega(J_{t+1},K_{t+1})|$ bins. The tosses are uniformly at random and independent of each other. Repeat this process $J_{t+1}$ times so that each ball $i$ belongs to $J_{t+1}$ distinct bins. Definitely, each bin contains $K_{t+1}$ balls on average. Once these balls and bins processes are completed, the collection of the corresponding subsets, i.e., 
\begin{equation}\label{eq:collection}
    \Omega(J_{t+1},K_{t+1})\eqdef \left\{\Vc_{i}\subseteq [P]:i=1,...,\left\lfloor \gamma_{t+1} P \right\rfloor\right\},
\end{equation} is formed such that $\Vc_{i}$ takes the balls' indices belong to the bin $i$ as elements. Also, the notation in (\ref{eq:collection}) can be rewritten as
\begin{equation}
    \Omega(K_{t+1}) = \Omega\left(J_{t+1}={ P \choose K_{t+1}}, K_{t+1}\right).
\end{equation} That is, with the particular choice of $J_{t+1}$, the above collection contains the all subsets of size $K_{t+1}$ as before. Also, we remark that the proposed collection $\Omega(J_{t+1}, K_{t+1})$ satisfies the uniform frequency, i.e., each kernel occurs exactly $J_{t+1}$ times.\hfill$\Diamond$
\end{remark}
\vspace{0.2cm}

Finally, we propose a randomized algorithm to choose a subset of kernels from $\Omega(J_{t+1},K_{t+1})$. Define a discrete random variable $S_{t+1}$ with the probability mass function (PMF):
\begin{equation}\label{eq:PMF1}
\hat{\alpha}_{t+1}(j)= \frac{\sum_{i \in \Vc_{j}}\hat{w}_{t+1}(i)}{J_{t+1} \sum_{i=1}^{P}\hat{w}_{t+1}(i)}, 
\end{equation} for $j \in \left[|\Omega(J_{t+1},K_{t+1})|\right]$, where $\hat{w}_{t+1}(i)$ is defined in (\ref{eq:o_w}). Due to the uniform frequency (see Remark~\ref{rem:RC}), we can easily verify that (\ref{eq:PMF1}) is a valid PMF. Letting
\begin{equation}\label{eq:PMF11}
    \hat{\alphav}_{t+1}=(\hat{\alpha}_{t+1}(1),...,\hat{\alpha}_{t+1}(|\Omega(J_{t+1},K_{t+1})|)),
\end{equation} OMKL-AKS chooses a subset in the following way:
\begin{itemize}
    \item Sampling $S_{t+1}$ according to $\hat{\alphav}_{t+1}$ in (\ref{eq:PMF11}).  The corresponding sample is denoted as $s_{t+1}$.
    \item Accordingly, the selected subset is denoted as $\Vc_{s_{t+1}} \in \Omega(J_{t+1}, K_{t+1})$.
\end{itemize} 
\begin{example}
Consider the example of $P=6$ and $\Omega(2,3)=\{\Vc_1=\{1,3,4\},\Vc_2=\{1,3,5\},\Vc_3=\{2,5,6\},\Vc_4=\{2,4,6\} \}$. When $s_t=3$, the subset of selected kernels is equal to $\Vc_{3}=\{2,5,6\}$.\hfill$\Diamond$
\end{example}

Given the selected subset $\Vc_{s_{t+1}}$, OMKL-ASK learns a target function $\hat{f}_{t+1}(\xv)$ as
\begin{equation}\label{eq:OMKL-AKS_f}
    \hat{f}_{t+1}(\xv) = \sum_{i \in \Vc_{s_{t+1}}} \hat{q}_{t+1}(i) \hat{f}_{t+1,i}(\xv).
\end{equation} where the weight distribution is given as
\begin{equation}\label{eq:weight_OMKL_AKS}
    \hat{q}_{t+1}(i) = \frac{\hat{w}_{t+1}(i)}{\sum_{\ell \in \Vc_{s_{t+1}}} \hat{w}_{t+1}(\ell)},
\end{equation} for $i \in \Vc_{s_{t+1}}$.  The specific procedures are provided in {\bf Algorithm 1}. We remark that OMKL-AKS includes the conventional OMKL (a.k.a., Raker) with the particular choices of subsets as $\Vc_{s_{t}}=[P]$ for all $t\in [T]$.

\begin{algorithm}[t]
\caption{OMKL-AKS }
\begin{algorithmic}[1]
\State {\bf Input:} Kernels $\kappa_i$, $i\in[P]$, parameters $\eta_l, \eta_g, \gamma_t>0$, the number of random features $D$ (for RF approximation). 
\State {\bf Output:} A sequence of functions $\hat{f}_{t}(\xv)$ for $t\in [T+1]$.
\vspace{0.05cm}
\State {\bf Initialization:}  $\hat{\thetav}_{1,i} = \zerov$ (i.e., $\hat{f}_{1,i}=0$), and $\hat{w}_{1}(i) = 1$ for $i\in[P]$.
\vspace{0.05cm}

\State {\bf Iteration:} $t=1,...,T$.

\hspace{-0.8cm}$\bullet$ Receive a labeled data $(\xv_t, y_t)$.

\hspace{-0.8cm}$\bullet$ Construct $\zv_{i}(\xv_t)$ via (\ref{eq:zv}) using the kernel $\kappa_i$ for $i\in[P]$.

\hspace{-0.8cm}$\bullet$ {\em Local step:}
\begin{itemize}
\item[$-$] Update  $\hat{\thetav}_{t+1,i}$ via OGD in (\ref{eq:OCO}).
\item[$-$]  Set $\hat{f}_{t+1,i}(\xv)=\hat{\thetav}_{t+1,i}\zv_{i}(\xv)$ for $i\in [P]$.
\end{itemize}

\hspace{-0.8cm}$\bullet$ {\em Global step:}
\begin{itemize}
    \item[$-$] Kernel selection:
    \begin{itemize}
    \item[$\circ$] Obtain $K_{t+1}$ via (\ref{eq:d_r_t}) and set $J_{t+1}=\gamma_{t+1} K_{t+1}$ 
    \item[$\circ$] Construct $\Omega(J_{t+1}, K_{t+1})$ from Remark~\ref{rem:RC}.
    \item[$\circ$] Obtain $\hat{\alphav}_{t+1}$ via (\ref{eq:PMF11}).
    \item[$\circ$] Choose a subset $\Vc_{s_{t+1}}\in \Omega(J_{t+1},K_{t+1})$ according to $S_{t+1} \sim \hat{\alphav}_{t+1}$. 
   \end{itemize}
    \item[$-$]  Update $\hat{w}_{t+1}(i)$ via (\ref{eq:o_w}) for $i\in[P]$.
    \item[$-$]  Obtain $\hat{q}_{t+1}(i)$ from (\ref{eq:weight_OMKL_AKS}) for $i\in \Vc_{s_t}$.
    \item[$-$]  Update $\hat{f}_{t+1}(\xv) = \sum_{i\in \Vc_{s_{t+1}}}\hat{q}_{t+1}(i)\hat{f}_{t+1,i}(\xv)$.
\end{itemize}
\end{algorithmic}
$\divideontimes$ RF-based OMKL (a.k.a., Raker) performs with $\Vc_{s_{t+1}}=[P]$ for all $t\in[T]$, where kernel selection in global step is skipped.
\end{algorithm}

\subsection{The Proposed AMKL-AKS}\label{subsec:AMKL}

We propose a streaming-based active learning for OMKL and OMKL-AKS frameworks in Section~\ref{subset:OMKL}, which are respectively called AMKL and AMKL-AKS. In OMKL frameworks, the label of every incoming data is always revealed to the learner. Whereas, in AMKL frameworks, the label of an incoming data is identified only when the learner has made a request to acquire the label from an oracle. This process is necessary in many real-world applications as the label acquisition can be expensive and time-consuming. Our goal is to construct AMKL algorithms which can achieve the almost same accuracy of OMKL counterparts with a smaller number of labeled samples. In these extensions, the key challenge is to decide when the learner should or should not acquire the label of an incoming data from the oracle. To perform this process efficiently, we develop a {\em selection criterion} suitable for OMKL frameworks in  Section~\ref{subset:OMKL}. Then, the proposed AMKL algorithms operate as follows: they skips the label request for an incoming data if the selection criterion is satisfied, and directly follow OMKL frameworks, otherwise. Obviously, the selection criterion plays a key role in determining the accuracy and efficiency of the proposed AMKL algorithms. To explain AMKL frameworks, we follow the notations, definitions, and procedures in Section~\ref{subset:OMKL}, and we in this section highlight the major differences.

We first introduce a binary variable $a_t \in \{0,1\}$, $t\in [T]$ to indicate the time indices of $y_t$ being revealed, i.e.,  $a_t=1$ if the learner requested the label of an incoming data $\xv_{t}$ (i.e., the selection criterion is not satisfied), and $a_t=0$, otherwise. At time $t$, the learner receives a new data $\xv_{t}$, and from the previous time slots,  the kernel functions  $\{\bar{f}_{t,i}(\xv): i\in [P]\}$,  weight distributions $\{\bar{p}_{t}(i): i\in [P]\}$ (also $\{\bar{w}_{t}(i): i\in[P]\}$), and a subset $\Vc_{s_{t}}$ are known. Clearly,  $\bar{f}_{t,i}$ and $\bar{p}_{t}(i)$ can be different from $\hat{f}_{t,i}$ and $\hat{p}_{t}(i)$ in OMKL algorithms, provided that at least one label is not revealed. Focusing on time $t$, the proposed AMKL-AKS algorithm proceeds with the following three steps. We notice that AMKL (without subset selection) exactly follows the same procedures of AMKL-AKS, with  $\Vc_{s_{t}}=[P]$ for all $t\in[T]$.

\vspace{0.1cm}
{\em i) Active labeling step:} This step decides whether or not to acquire the label of an incoming data $\xv_t$ from the oracle, where the decision is made by the proposed selection criterion below. Regarding the selection criterion, we first propose a confidence condition as
\begin{equation}
\max_{j\in [P]} \sum_{i\in \Vc_{s_{t}}} \bar{p}_{t}(i)\Lc(\bar{f}_{t,i}(\xv_{t}),\bar{f}_{t,j}(\xv_{t})) \leq \eta_{c},\label{eq:confidence_a}
\end{equation} for some parameter $\eta_c>0$. As seen in (\ref{eq:confidence_a}), this condition is simply checked with the current local functions, without knowing the label of the incoming data. For AMKL (without a subset selection), the above condition is slightly modified by setting $\Vc_{s_{t}}=[P]$. Intuitively, the confidence condition in (\ref{eq:confidence_a}) can ensure that the accuracy difference from OMKL counterpart, obtained using the true label $y_t$, can be bounded by a small value $\eta_c$, where $\eta_c$ is determined by considering an accuracy-efficiency tradeoff. In other words, the labeling does not improve the weight distributions, i.e., $\hat{\pv}_{t+1} \approx \bar{\pv}_{t+1}$. The theoretical evidence is provided in Lemma~\ref{lem4}. Thus, if the confidence condition in (\ref{eq:confidence_a} holds, it would be better to skip the label-request in terms of the accuracy-efficiency tradeoff. It is remarkable that skipping the label-request also impacts on the updates of kernel functions (i.e., OGD updates) as well as the update of weight distributions. Thus, we introduce a parameter $M$ to ensure a sufficient local updates (i.e., OGD updates), where $M$ represents the maximum number of consecutive unlabeling data. In Lemma~\ref{lem3}, it is proved that OGD updates achieve an optimal sublinear regret as long as $M$ is a constant (i.e., does not grow with $T$). In non-asymptotic cases, $M$ can be chosen by taking into account the accuracy-efficiency tradeoff. To sum up, the {\em selection criterion} for a labeling request (i.e., to determine the indicative variable $a_t\in\{0,1\}$) is proposed as 

{\bf (Selection criterion)} The label of an incoming data $x_t$ is revealed (i.e., $a_t=1$) only when $\sum_{\tau=1}^{M}a_{t-\tau}\neq 0$ and the confidence condition in (\ref{eq:confidence_a}) is satisfied.

It is theoretically proved that the proposed selection criterion can keep the optimal sublinear regret, implying that it indeed avoids unuseful label-requests. Given a sequence of indicate variables $\{a_t: t\in [T]\}$, the active-learning efficiency is defined as 
\begin{equation}\label{eq:eff}
    {\rm AL_{eff}} \eqdef \frac{\sum_{i=1}^T a(t)}{T}.
\end{equation} Given the parameter $M$, AL efficiency is lower-bounded as
\begin{equation}
    {\rm AL_{eff}} \geq 1 - \frac{M}{M+1}.
\end{equation}

\begin{algorithm}
\caption{AMKL-AKS}
\begin{algorithmic}[1]
\State {\bf Input:} Kernels $\kappa_i$, $i\in[P]$, parameters $\eta_l, \eta_g, \eta_{c}, \gamma_t > 0, M\geq 1$,  the number of random features $D$ (for RF approximation).  
\State {\bf Output:} A sequence of functions $\bar{f}_{t}(\xv)$, $t\in [T+1]$.
\vspace{0.05cm}
\State {\bf Initialization:}  $\bar{\thetav}_{1,i} = \zerov$ (i.e., $\bar{f}_{1,i}=0$), and $\bar{w}_{1}(i) = 1$ for $i\in[P]$.
\vspace{0.05cm}
\State {\bf Iteration:} $t=1,...,T$

\hspace{-0.8cm}$\bullet$ Receive a streaming data $\xv_t$.

\hspace{-0.8cm}$\bullet$ Construct $\zv_{i}(\xv_t)$ via (\ref{eq:zv}) using the kernel $\kappa_i$ for $i\in[P]$.

\hspace{-0.8cm}$\bullet$ {\em Active labeling step:} 
\begin{itemize}
    \item[$-$] If $\sum_{\tau=1}^{M}a_{t-\tau}\neq0$ and the confidence condition in (\ref{eq:confidence_a}) is satisfied:
    \begin{itemize}
        \item[$\circ$] Set $a_t = 0$ and $\bar{f}_{t+1}(\xv) = \bar{f}_{t}(\xv)$.
        \item[$\circ$] Skip the active local and global steps.
    \end{itemize}
    \item[$-$] Otherwise, set $a_t = 1$ and receive $y_t$ from the oracle.
\end{itemize}

\hspace{-0.8cm}$\bullet$ {\em Active local step ($a_t=1$):} 
\begin{itemize}
   \item[$-$] Update  $\bar{\thetav}_{t+1,i}$ via OGD in (\ref{eq:OCO}).
   \item[$-$] Set $\bar{f}_{t+1,i}(\xv)=\bar{\thetav}_{t+1,i}\zv_{i}(\xv)$ for $i\in [P]$.
\end{itemize}

\hspace{-0.8cm}$\bullet$ {\em Active global step ($a_t=1$):}
\begin{itemize}
    \item[$-$] Kernel selection:
    \begin{itemize}
        \item[$\circ$] Obtain $K_{t+1}$ via (\ref{eq:d_r_t_1}) and $J_{t+1}=\gamma_{t+1} K_{t+1}$.
        \item[$\circ$] Construct $\Omega(J_{t+1}, K_{t+1})$ from Remark~\ref{rem:RC}.
        \item[$\circ$] Obtain $\bar{\alphav}_{t+1}$ via (\ref{eq:PMF22}).
        \item[$\circ$] Choose a subset $\Vc_{s_{t+1}}\in \Omega(J_{t+1},K_{t+1})$ according to PMF $S_{t+1} \sim \bar{\alphav}_{t+1}$.
    \end{itemize}
    \item[$-$]  Update $\bar{w}_{t+1}(i)$ via (\ref{eq:a_w}).
    \item[$-$]  Obtain $\bar{q}_{t+1}(i)$ from (\ref{eq:weight_OMKL_AKS2}), for $i\in [P]$.
    \item[$-$] Update $\bar{f}_{t+1}(\xv) = \sum_{i\in \Vc_{s_{t+1}}}\bar{q}_{t+1}(i)\bar{f}_{t+1,i}(\xv)$.
    \end{itemize}
\end{algorithmic}
$\divideontimes$ AMKL performs with $\Vc_{s_{t+1}}=[P]$ for all $t\in[T]$, where kernel selection in active global step is skipped.
\end{algorithm}

\vspace{0.1cm}
{\em ii) Active local step:} Unlike OMKL frameworks, OGD update in (\ref{eq:OCO}) cannot proceed  when the label $y_{t}$ is not revealed (i.e., $a_{t}=0$). This is because in this case the loss function  $\Lc(\cdot, y_{t})$ is undefined. Accordingly, each kernel function $i$ in AMKL frameworks is optimized such as
\begin{equation}\label{eq:a_f}
    \bar{f}_{t+1,i}(\xv)=\bar{\thetav}_{t+1,i}^{\trasp}\zv_{i}(\xv),\;\; i\in[P],
\end{equation} where
\begin{equation}\label{eq:a_online}
    \bar{\thetav}_{t+1,i} = 
    \begin{cases}
    \bar{\thetav}_{t,i},\; \mbox{ if }a_{t}=0\\
    \bar{\thetav}_{t}  - \eta_{l}\nabla\Lc(\hat{\thetav}_{t,i}^{\trasp}\zv_{i}(\xv_{t},y_{t})),\; \mbox{ if } a_{t}=1.
    \end{cases}
\end{equation} That is, when $a_{t}=0$, kernel functions are not updated as $\bar{f}_{t+1,i}(\xv) = \bar{f}_{t,i}(\xv)$ for $i\in[P]$.

\vspace{0.1cm}
{\em iii) Active global step:} In this step, the weights should be modified since the loss values of unlabeled samples are not revealed. Thus, they are computed as
\begin{equation} \label{eq:a_p}
    \bar{p}_{t+1}(i) = \frac{\bar{w}_{t+1}(i)}{\sum_{i=1}^{P}\bar{w}_{t+1}(i)},\;\; i\in[P],
\end{equation} where 
\begin{equation} \label{eq:a_w}
\bar{w}_{t+1}(i)=\exp\left(-\eta_{g}\sum_{\tau=1}^{t}a_{\tau}\Lc(\hat{f}_{i,\tau}(\xv_\tau),y_\tau)\right).
\end{equation} Note $\bar{w}_{t+1}(i)\neq \hat{w}_{t+1}(i)$, provided that at least one label is not revealed during the previous time slots. As in OMKL-AKS, the parameters of an adaptive kernel selection determined as
\begin{equation}\label{eq:d_r_t_1}
    K_{t+1} = \left|\left\{i\in[P]: \bar{p}_{t+1}(i)/\bar{p}^{\star}_{t+1} > \delta_{t+1} \right\}\right|,
\end{equation}  and $J_{t+1}=\gamma_{t+1}K_{t+1}$. Also, given the $J_{t+1}$ and $K_{t+1}$, define a kernel-selection probability distribution:
\begin{equation}\label{eq:PMF2}
\bar{\alpha}_{t+1}(j)= \frac{\sum_{i \in \Vc_{j}}\bar{w}_{t+1}(i)}{d_{l,t+1} \sum_{i=1}^{P}\bar{w}_{t+1}(i)},
\end{equation} for $j \in [|\Omega(d_{l,t+1},d_{r,t+1})|]$, where $\bar{w}_{t+1}(i)$ is defined in (\ref{eq:a_w}). Letting
\begin{equation}\label{eq:PMF22}
    \bar{\alphav}_{t+1} = (\bar{\alpha}_{t+1}(1),...,\bar{\alpha}_{t+1}(|\Omega(J_{t+1},K_{t+1})|)),
\end{equation}AMKL-AKS selects the subset to be used at time $t$ as follows:
\begin{itemize}
    \item Sampling $S_{t+1}$ according to $\bar{\alphav}_{t+1}$ in (\ref{eq:PMF22}). 
    \item Then, the chosen subset at time $t$ is given as $\Vc_{s_{t+1}} \in \Omega(J_{t+1}, K_{t+1})$.
\end{itemize}Given the subset $\Vc_{s_{t+1}}$, AMKL-ASK learns a target function $\bar{f}_{t+1}(\xv)$ as
\begin{equation}\label{eq:OMKL-AKS_f2}
    \bar{f}_{t+1}(\xv) = \sum_{i \in \Vc_{s_{t+1}}} \bar{q}_{t+1}(i) \bar{f}_{t+1,i}(\xv).
\end{equation} where the weight distribution is given as
\begin{equation}\label{eq:weight_OMKL_AKS2}
    \bar{q}_{t+1}(i) = \frac{\bar{w}_{t+1}(i)}{\sum_{\ell \in \Vc_{s_{t+1}}} \bar{w}_{t+1}(\ell) },
\end{equation} for $i \in \Vc_{s_{t+1}}$. The specific procedures are provided in {\bf Algorithm 2}.


%
%

\section{Regret Analysis}\label{sec:regret}

We analyze the cumulative regrets of the proposed online and active learning algorithms. For the regret analysis of this section, the following conditions are assumed:
\begin{itemize}
    \item {\bf (a1)} For any fixed $\zv_{i}(\xv_t)$ and $y_t$, the loss function $\Lc(\thetav^{\trasp}\zv_{i}(\xv_{t}), y_t)= \Lc(y_t, \thetav^{\trasp}\zv_{i}(\xv_{t}))$ is convex with respect to $\thetav$, and is bounded as $\Lc(\thetav^{\trasp}\zv_{i}(\xv_t),y_t) \in [0,\ell_u]$.
    \item {\bf (a2)} For any kernel $i$, $\thetav_{t,i}$ belongs to a bounded set $\Theta_i\subseteq \RR^{2D}$, i.e., $\|\thetav_{t_1,i} - \thetav_{t_2,i}\| \leq C$ for any $t_1,t_2 \in [T]$.
    \item {\bf (a3)} The loss function is $L$-Lipschitz continuous, i.e., $\|\nabla \Lc(\thetav^{\trasp}\zv_{i}(\xv_t),y_t)\| \leq L$.
\end{itemize} It is remarkable that (a1)-(a3) are usually assumed for the analysis of  online convex optimizations and online learning frameworks \cite{rahimi2008random,bubeck2011introduction,shen2019random}. Also, let $f_{i}^{\star}(\xv)=(\thetav_{i}^{\star})^{\trasp}\zv_i(\xv)$ denote the optimal RF approximation function at the kernel $i$, i.e.,
\begin{equation}
    \thetav_{i}^{\star} \eqdef \arg\min_{\thetav \in \Theta_i} \sum_{t=1}^{T} \Lc\left(\thetav^{\trasp}\zv_{i}(\xv_t), y_t\right),\;\; i \in[P].
\end{equation}

We state the main results of this section, i.e., the regret analysis of the proposed OMKL-AKS, AMKL, and AMKL-AKS.

\vspace{0.1cm}
\begin{theorem}\label{thm2} For any small $\delta>0$, OMKL-AKS with parameters $\eta_{l}=\eta_{g}=\mathcal{O}(1/\sqrt{T})$ guarantees the following regret bound with probability 1-$\delta$:
\begin{align*}
&{\rm regret}_{T}^{\rm OL-A} \\
&= \sum_{t=1}^{T}\Lc\left(\sum_{i \in \Vc_{S_t}} \hat{q}_{t}(i) \hat{f}_{t,i}(\xv_t),y_{t}\right) - \min_{1\leq i \leq P}\sum_{t=1}^{T}\mathcal{L}\left(f^{\star}_{i}({\bf x}_{t}),y_{t}\right)\\
&\leq \mathcal{O}(\sqrt{T}),
\end{align*} where a randomness is from an internal random kernel selection.\hfill$\blacksquare$
\end{theorem}
\begin{remark}
We emphasize that Theorem~\ref{thm2} is valid with any choices of $J_{t}$ and $K_{t}$ as long as the uniform frequency in the construction of collection (i.e., set of subsets of $P$ kernels) is satisfied, i.e., each kernel $i$ occurs exactly $J_t$ times in the collection. Note that OMKL-AKS with $K_t = P$ for all $t \in [T]$ is equivalent to OMKL (a.k.a., Raker). In this case, the analysis in  Theorem~\ref{thm2} holds with $\delta=0$ as the randomness for a random subset selection disappears. Thus, Theorem~\ref{thm2} encompasses the regret analysis in \cite{shen2019random}. \hfill$\Diamond$
\end{remark}
\vspace{0.2cm}

For the analysis of active learning, we further assume that
\begin{itemize}
    \item {\bf (a4)} If $\Lc(\bar{\thetav}_{t,i}^{\trasp}\uv_t, \bar{\thetav}_{t,j}^{\trasp}\uv_t)\leq \epsilon$ for an input $\uv_t$ with $\|\uv_t\|=1$, then there exists a small $B>0$ such that $\Lc(\bar{\thetav}_{t,i}^{\trasp}\uv, \bar{\thetav}_{t,j}^{\trasp}\uv)\leq \epsilon B$ for any $\uv$ with $\|\uv\|=1$.
    \item {\bf (a5)} $\Lc(\cdot,\cdot)$ obeys the triangle inequality.
\end{itemize}  For example, 0-1 loss for classification and $\ell_1$/$\ell_2$-norm loss in regression satisfy the triangle inequality.

\vspace{0.1cm}
\begin{theorem}\label{thm3}
 AMKL with the parameters $\eta_l=\eta_g=\eta_c=\mathcal{O}(1/\sqrt{T})$ guarantees the sublinear regret as
\begin{align*}
    &{\rm regret}_{T}^{\rm AL}\\
    & = \sum_{t=1}^{T}\Lc\left(\bar{f}_{t}(\xv_t), y_t\right) - \min_{1\leq i \leq P}\Lc\left(f_{i}^{\star}(\xv_t),y_t\right)\leq \mathcal{O}(\sqrt{T}).
\end{align*} \hfill$\blacksquare$
\end{theorem}

\vspace{0.1cm}
\begin{theorem}\label{thm4}
For any $\delta>0$, AMKL-AKS  with the parameters $\eta_l=\eta_g=\eta_c=\mathcal{O}(1/\sqrt{T})$ guarantees the sublinear regret with probability $1-\delta$ as
\begin{align*}
    &{\rm regret}_{T}^{\rm AL-A}\\
    &= \sum_{t=1}^{T}\Lc\left(\sum_{i \in \Vc_{S_t}} \bar{q}_{t}(i) \bar{f}_{t,i}(\xv_t),y_{t}\right) - \min_{1\leq i \leq P}\Lc\left(f_{i}^{\star}(\xv_t),y_t\right)\\
    &\leq \mathcal{O}(\sqrt{T}).
\end{align*} \hfill$\blacksquare$
\end{theorem} 
\vspace{0.2cm}

The proofs of the main theorems will be provided in the following subsections.

\subsection{Proof of Theorem~\ref{thm2}}

We prove that the proposed OMKL-AKS achieves the sublinear regret with high probability. We provide key lemmas for the proof of Theorem~\ref{thm2}.  Lemma~\ref{lem1} below states that OGD (online-gradient descent) in local update can guarantee the sublinear regret.

\begin{lemma}\label{lem1} For any kernel $i$, online gradient descent in (\ref{eq:OCO}) with step size $\eta_{l}$ guarantees the following:
\begin{align*}
    {\rm regret}_{T}^{{\rm l}} &=\sum_{t=1}^{T}\mathcal{L}\left(\hat{f}_{t,i} ({\bf x}_{t}),y_{t}\right)- \sum_{t=1}^{T} \mathcal{L}\left(f_{i}^{\star}({\bf x}_{t}),y_{t}\right)\\
    &\leq \frac{C^{2}}{2\eta_{l}}+\frac{\eta_{l}L^{2}T}{2}.
\end{align*}
\end{lemma}
\begin{IEEEproof}
    The proof is immediately done from the proof of \cite[Theorem 3.1]{hazan2016introduction} since $\Lc(\cdot,y_t)$ is a convex function for a fixed label $y_t$.
\end{IEEEproof}

The following lemma shows that OMKL with EXP strategy can ensure the sublinear regret over the best kernel performance.
\begin{lemma}\label{lem2} For any fixed $\eta_{g}>0$, OMKL with the EXP strategy in (\ref{eq:exp_st}) satisfies 
\begin{align*}
{\rm regret}_{T}^{\rm g}& = \sum_{t=1}^{T} \sum_{i=1}^{P} \hat{p}_{t}(i) \mathcal{L}\left(\hat{f}_{t,i} ({\bf x}_{t}),y_{t}\right) \\
&- \min_{1\leq i \leq P}\sum_{t=1}^{T}\mathcal{L}\left({\hat{f}_{t,i}}({\bf x}_{t}),y_{t}\right) \leq \frac{\log{P}}{\eta_g} + \frac{\eta_g T \ell_{u}^2}{8}.
\end{align*}
\end{lemma}
\begin{IEEEproof}
The proof is provided in Appendix~\ref{app:lem2}.
\end{IEEEproof}
\vspace{0.1cm}

We are now ready to prove Theorem~\ref{thm2}.  Note that Lemma~\ref{lem1} holds for any kernel $i$. Thus, from Lemma~\ref{lem1}, Lemma~\ref{lem2}, and the convexity of the loss function $\Lc(\cdot,y_t)$ for any fixed label $y_t$, we can get:
\begin{align}
&{\rm regret}_{T}^{{\rm OL}} \nonumber\\
&= \sum_{t=1}^{T}\mathcal{L}\left(\sum_{i=1}^{P}\hat{p}_{t}(i)\hat{f}_{t,i} (\xv_{t}),y_{t}\right)- \min_{1\leq i \leq P}\sum_{t=1}^{T}\mathcal{L}\left(f^{\star}_{i}({\bf x}_{t}),y_{t}\right)\nonumber\\
&\leq \frac{C^{2}}{2\eta_{l}}+\frac{\eta_{l}{L}^{2}T}{2}+\frac{\log{P}}{\eta_g} + \frac{\eta_g T\ell_{u}^2}{8}.\label{eq:sublinear_OMKL}
\end{align} Setting $\eta_{l} = \frac{C}{\sqrt{T}}$ and $\eta_{g} = 2\sqrt{\frac{2\log{P}}{T}}$, OMKL (or Raker) guarantees the sublinear regret $\mathcal{O}(\sqrt{T})$. This proves the special case of Theorem~\ref{thm2} with $K_t=P$ for all $t\in[T]$. Then, the general case will be proved using Azuma-Hoeffding's ineqality (i.e., the concentration bound for a martingale difference sequence). Define a random variable $X_{t}$ as
\begin{align*}
    X_{t} &= \sum_{i \in \Vc_{S_t}}\frac{\hat{w}_{t}(i)}{\sum_{\ell \in \Vc_{S_t}}\hat{w}_{t}(l)} \Lc(\hat{f}_{t,i}(\xv_t),y_t) - U_t,
\end{align*} where 
\begin{equation}
    U_t = \sum_{i=1}^{P}\hat{p}_{t}(i)\Lc(\hat{f}_{t,i}(\xv_t),y_t).
\end{equation}
Let $\Fc_{t}=\sigma(S_{1},S_{2},...,S_{t})$ be the smallest signal algebra such that $S_1,S_2,...,S_t$ is measurable. Then, $\left\{\Fc_{t}:t=1,...,T\right\}$ is filtration and $X_t$ is $\Fc_t$ measurable.Note that condition on $\Fc_{t-1}$, the $\hat{w}_{t}(i)$ in (\ref{eq:o_w}), $\hat{p}_{t-1}(i)$ in (\ref{eq:exp_st}), and $q_{S_{t}}(i)$ in (\ref{eq:PMF1}) are fixed, and $S_{t}$ is  only random variable. Using this fact, we first show that $\{X_1,...,X_T\}$ is a martingale difference sequence with respect to filtration $\Fc_1\subseteq \Fc_{2}\subseteq \cdots \subseteq \Fc_{T}$, by showing that
\begin{equation}\label{eq:claim1}
    \EE\left[X_t|\Fc_{t-1}\right]=0.
\end{equation}  Then, this claim is proved as follows:
\begin{align*}
    &\EE\left[X_t|\Fc_{t-1}\right]\\
    &=\EE\left[\sum_{i \in \Vc_{S_t}}\frac{\hat{w}_{t}(i)}{\sum_{\ell \in \Vc_{S_t}}\hat{w}_{t}(\ell)} \Lc(\hat{f}_{t,i}(\xv_t),y_t)-U_t\Big|\Fc_{t-1}\right]\\
    &\stackrel{(a)}{=}\EE\left[\sum_{i \in \Vc_{S_t}}\frac{\hat{w}_{t}(i)}{\sum_{\ell \in \Vc_{S_t}}\hat{w}_{t}(\ell)} \Lc(\hat{f}_{t,i}(\xv_t),y_t)\Big|\Fc_{t-1}\right]-U_t
\end{align*}
\begin{align*}
    &\stackrel{(b)}{=}\sum_{j=1}^{|\Omega(J_{t},K_{t})|} q_{S_t}(j)\left( \sum_{i \in \Vc_{j}}\frac{\hat{w}_{t}(i)}{\sum_{\ell \in \Vc_{j}}\hat{w}_{t}(\ell)} \Lc(\hat{f}_{t,i}(\xv_t),y_t)\right)-U_t\\
     &= \sum_{j=1}^{|\Omega(J_{t},K_{t})|}\frac{\sum_{i \in \Vc_{j}}\hat{w}_{t}(i)}{J_{t}\sum_{i=1}^{P}\hat{w}_{t}(i)}\Lc(\hat{f}_{t,i}(\xv_t),y_t)-U_t\\
    &\stackrel{(c)}{=}0,
\end{align*}where (a) and (b) follow from the fact that $\hat{w}_{t}(i)$, $\hat{p}_t(i)$, $q_{S_t}(i)$ are functions of random variables $S_{1},...,S_{t-1}$, and $(c)$ follows the
\begin{equation*}
    \sum_{j=1}^{|\Omega(J_{t},K_{t})|}\sum_{i \in \Vc_{j}}\hat{w}_{t}(i)\Lc(\hat{f}_{t,i}(\xv_t),y_t) = J_{t}\sum_{i=1}^{P}\hat{w}_{t}(i)\Lc(\hat{f}_{t,i}(\xv_t),y_t).
\end{equation*} Since $\{X_t: t\in[T]\}$ is a martingale difference sequence and $X_t \in [A_t,A_t+c_t]$ is bounded, where 
\begin{equation}
    A_t=-\sum_{i=1}^{P}\hat{p}_{t}(i)\Lc(\hat{f}_{t,i}(\xv_t),y_t),
\end{equation} is a random variable and $\Fc_{t-1}$ measurable, and $c_t = \ell_u$. From Azuma-Hoeffding's inequality \cite{wainwright2019high}, the following bound holds for some $\delta>0$ with high probability $1-\delta$:
\begin{align}
    \sum_{t=1}^{T} X_{t}&= \sum_{t=1}^{T}\sum_{i \in \Vc_{S_t}}\frac{\hat{w}_{t}(i)}{\sum_{l \in \Vc_{S_t}}\hat{w}_{t}(l)} \Lc(\hat{f}_{t,i}(\xv_t),y_t)\nonumber\\ & -\sum_{t=1}^{T}\sum_{i=1}^{P}\hat{p}_{t}(i)\Lc(\hat{f}_{t,i}(\xv_t),y_t) \nonumber\\
    &\leq \sqrt{\frac{\log(\delta^{-1})}{2}T\ell_{u}^2}.\label{eq:1}
\end{align} From (\ref{eq:1}), the following bound holds with probability $1-\delta$:
\begin{align*}
    &\sum_{t=1}^{T}\sum_{i \in \Vc_{S_t}}\frac{\hat{w}_{t}(i)}{\sum_{\ell \in \Vc_{S_t}}\hat{w}_{t}(\ell)} \Lc(\hat{f}_{t,i}(\xv_t),y_t) - \min_{1\leq i \leq P}\sum_{t=1}^{T}\Lc\left(f^{\star}_{i}(\xv_{t}),y_{t}\right) \nonumber\\
    &\leq \sum_{t=1}^{T}\sum_{i=1}^{P}\hat{p}_{t}(i)\Lc(\hat{f}_{t,i}(\xv_t),y_t) - \min_{1\leq i \leq P}\sum_{t=1}^{T}\Lc\left(f^{\star}_{i}(\xv_{t}),y_{t}\right)\\
    &+ \sqrt{\frac{T\ell_{u}^2\log(\delta^{-1})}{2}}\nonumber\\
    &\stackrel{(a)}{\leq} \frac{C^{2}}{2\eta_{l}}+\frac{\eta_{l}{L}^{2}T}{2}+\frac{\log{P}}{\eta_g} + \frac{\eta_g T\ell_{u}^2}{8} + \ell_{u}\sqrt{\frac{T\log(\delta^{-1})}{2}},
\end{align*} where (a) directly follows from Lemma~\ref{lem1} and Lemma~\ref{lem2}. The proof is completed from the convexity of the loss function and by setting $\eta_l=\eta_g=\mathcal{O}(1\sqrt{T})$.

\begin{table*}[!t]
\caption{ Summary of real datasets used in experiments.}
\label{tb:DataSummary}
\centering
\begin{tabular}{ c||c|c|c } 
\hline
Data sets & \# of features & \# of samples (T) & feature type \\
\hline
Twitter  & 77 & 13818 & real \& integer  \\ 
\hline
Twitter (Large)  & 77 & 98704 & real\& integer \\ 
\hline
Tom's hardware  & 96 & 9725 & real\& integer \\ 
\hline
Air quality  & 13 & 7322 & real \\ 
\hline
Appliance energy  & 25 & 18604 & real \\ 
\hline
Naval propulsion plants  & 16 & 11934 & real \\ 
\hline
\end{tabular}
\end{table*}

\subsection{Proofs of Theorem~\ref{thm3} and Theorem~\ref{thm4}}

We prove the optimal sublinear regrets of the proposed AMKL and AMKL-AKS.   We first derive the regret analysis of OGD in the active local step. This analysis is different from Lemma~\ref{lem1} since as shown in (\ref{eq:a_online}),  kernel functions cannot be updated at some times. The following lemma shows that the active local step can still guarantee the sublinear regret as long as the number of consecutive unlabeled samples is a certain constant (i.e., not grow with $T$).

\begin{lemma}\label{lem3} Let $M$ denote the maximum consecutive zeros (i.e., unlabeling) in $\{a_t: t\in[T]\}$. For any kernel $i$, online gradient descent in (\ref{eq:a_online}) (i.e., in active local step) with step size $\eta_{\ell}$ guarantees the following:
\begin{align*}
    {\rm regret}_{T}^{{\rm al}} &=\sum_{t=1}^{T}\mathcal{L}\left(\bar{f}_{t,i} ({\bf x}_{t}),y_{t}\right)- \sum_{t=1}^{T} \mathcal{L}\left(f_{i}^{\star}({\bf x}_{t}),y_{t}\right)\\ &\leq \frac{M+1}{2}\left(\frac{C^{2}}{\eta_{l}}+ \eta_{l}L^{2}T\right).
\end{align*} Setting $\eta_l=\mathcal{O}(1/\sqrt{T})$, OGD in the active local step can achieve the sublinear regret.
\end{lemma}
\begin{IEEEproof}
The proof is provided in Appendix~\ref{app:lem3}.
\end{IEEEproof}

\vspace{0.2cm}
For the purpose of AMKL analysis, we introduce a {\em virtual} OMKL. This method employs the same kernel functions with AMKL, i.e., both AMKL and virtual OMKL use the kernel functions $\bar{f}_{t,i}$, $i\in[P], t\in [T]$ in active local update. Whereas, in virtual OMKL, the weights are updated as if all labels $\{y_t: t\in [T]\}$ are revealed, namely,  
\begin{equation}\label{eq:v_exp_st}
    \tilde{p}_{t}(i) \eqdef \frac{\tilde{w}_{t}(i)}{\sum_{i=1}^{P}\tilde{w}_{t}(i)},
\end{equation} where 
\begin{equation} \label{eq:v_w}
\tilde{w}_{t}(i)=\exp\left(-\eta_{g}\sum_{\ell=1}^{t-1}\Lc(\bar{f}_{i,\ell}(\xv_\ell),y_\ell)\right),
\end{equation} for some parameter $\eta_g > 0$ and with the initial values $\tilde{w}_{1}(i)=1$, $i\in [P]$. Then, virtual OMKL learns a target function $\tilde{f}_t$ as
\begin{equation}
    \tilde{f}_{t}(\xv) = \sum_{i=1}^{P}\tilde{p}_t(i)\bar{f}_{t,i}(\xv).
\end{equation}
Comparing virtual OMKL and AMKL, we derive the following key lemmas.

\begin{figure*}[t]
\centering
\subfigure[Active-learning efficiency]{
\includegraphics[width=0.48\linewidth]{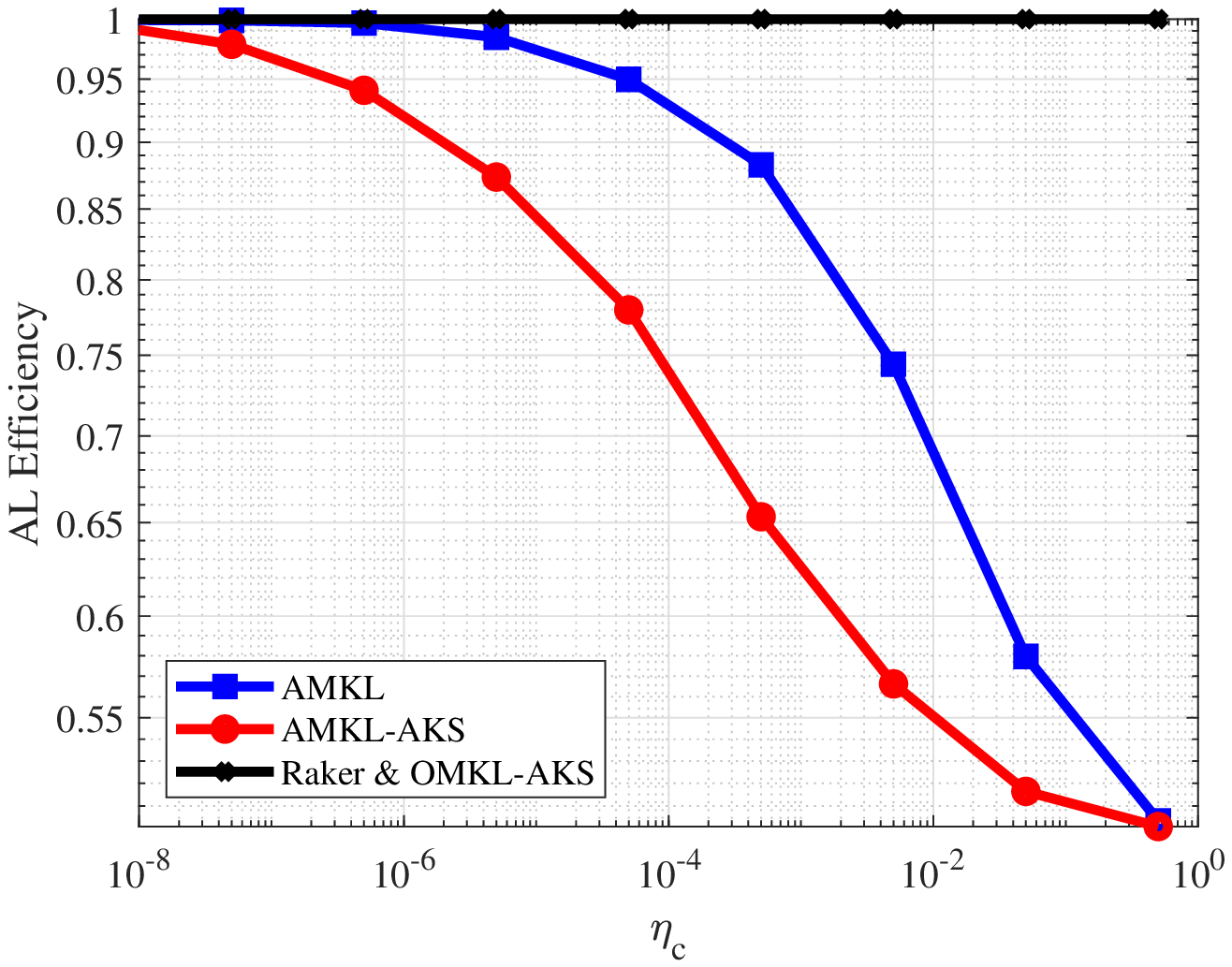}
}
\centering
\subfigure[MSE performance]{
\includegraphics[width=0.48\linewidth]{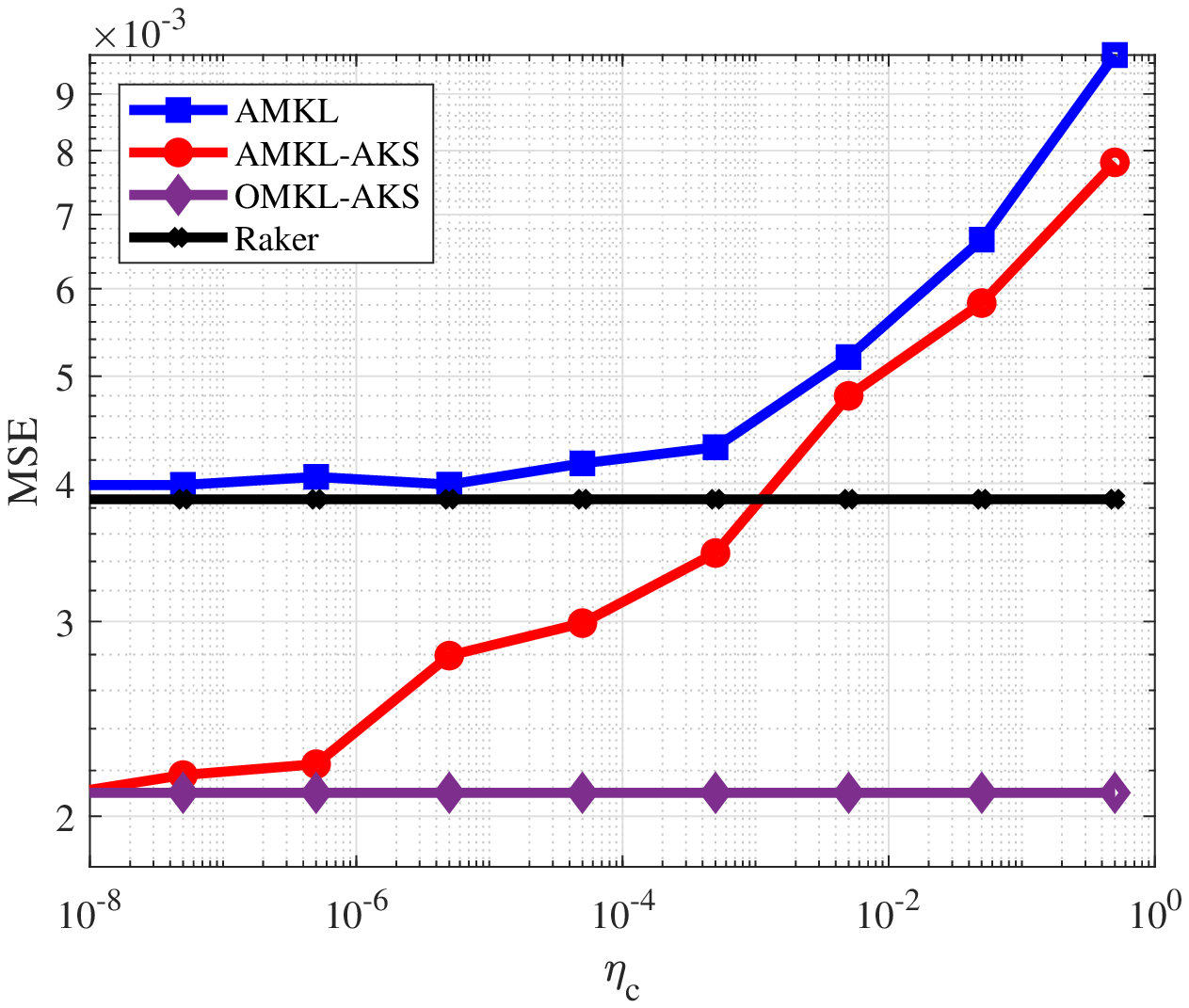}
}
\caption{Tradeoff between accuracy and efficiency of AMKL and AMKL-AKS for Tom's hardware dataset.} 
\label{fig:Eff-Acc}
\end{figure*}

\vspace{0.1cm}
\begin{lemma}\label{lem4}
For a small constant $\eta_c>0$ the confidence condition in (\ref{eq:confidence_a}) for AMKL (i.e., $\Vc_{s_t} = [P], t\in[T]$) implies
\begin{equation*}
    \Lc(\bar{f}_t(\xv_t),\tilde{f}_t(\xv_t))\leq \eta_c.
\end{equation*}
\end{lemma}
\begin{IEEEproof}
The proof is provided in Appendix~\ref{app:lem4}.
\end{IEEEproof}

\vspace{0.1cm}
\begin{lemma}\label{lem5} Letting $a_t = 0$ and $a_{t+1}\neq0$, the following inequality holds:
\begin{equation*}
    \Lc(\bar{f}_{t+1}(\xv_{t+1}), \tilde{f}_{t+1}(\xv_{t+1}))\leq \eta_c B.
\end{equation*}
\end{lemma}
\begin{IEEEproof}
The proof is provided in Appendix~\ref{app:lem4}.
\end{IEEEproof}

\vspace{0.1cm}
\begin{lemma}\label{lem6} Setting $\eta_c = \mathcal{O}(\sqrt{T})$, the following sublinear regret holds:
\begin{equation*}
    {\rm regret}_{T}^{\rm a}=\sum_{t=1}^{T}\Lc(\bar{f}_{t}(\xv_{t}), y_t) - \sum_{t=1}^{T} \Lc(\tilde{f}_{t}(\xv_{t}, y_t) \leq \mathcal{O}(\sqrt{T}).
\end{equation*}
\end{lemma}
\begin{IEEEproof}
The proof is provided in Appendix~\ref{app:lem6}.
\end{IEEEproof}

\vspace{0.2cm}
From now on, we will prove the main theorem using the above key lemmas.  From Lemma~\ref{lem2} and the convexity of the loss function, we can obtain the regret bound of vOMKL with $\eta_g = \mathcal{O}(1/\sqrt{T})$, which is given as
\begin{align*}
    {\rm regret}_T^{1} &= \sum_{t=1}^{T}\Lc(\tilde{f}_{t}(\xv_t),y_t) - \min_{1\leq i\leq P} \sum_{t=1}^{T} \Lc(\bar{f}_{i,t}(\xv_t),y_t) \\
    &\leq \mathcal{O}(\sqrt{T}).
\end{align*} Then, the proof is completed as
\begin{equation*}
    {\rm regret}_{T}^{\rm AL} = {\rm regret}_{T}^{\rm a} + {\rm regret}_{T}^{1} + {\rm regret}_{T}^{\rm al} \leq \mathcal{O}(\sqrt{T}),
\end{equation*} where ${\rm regret}_{T}^{\rm a}$ from Lemma~\ref{lem5} and ${\rm regret}_{T}^{\rm al}\leq \mathcal{O}(\sqrt{T})$ from Lemma~\ref{lem3} with $\eta_{l}=\mathcal{O}(1/\sqrt{T})$. This completes the proof of Theorem~\ref{thm3}. Also, in the proof of Theorem~\ref{thm2}, it was shown that the proposed kernel selection can keep the sublinear regret with high probability, as long as the underlying OMKL can do it. The same argument can be applied for the case of AMKL and AMKL-AKS. From  Theorem~\ref{thm2} and Theorem~\ref{thm3}, thus, the proof of Theorem~\ref{thm4} is completed.

\begin{figure*}[!h]
\centering
\subfigure[Twitter data]{
\includegraphics[width=0.45\linewidth]{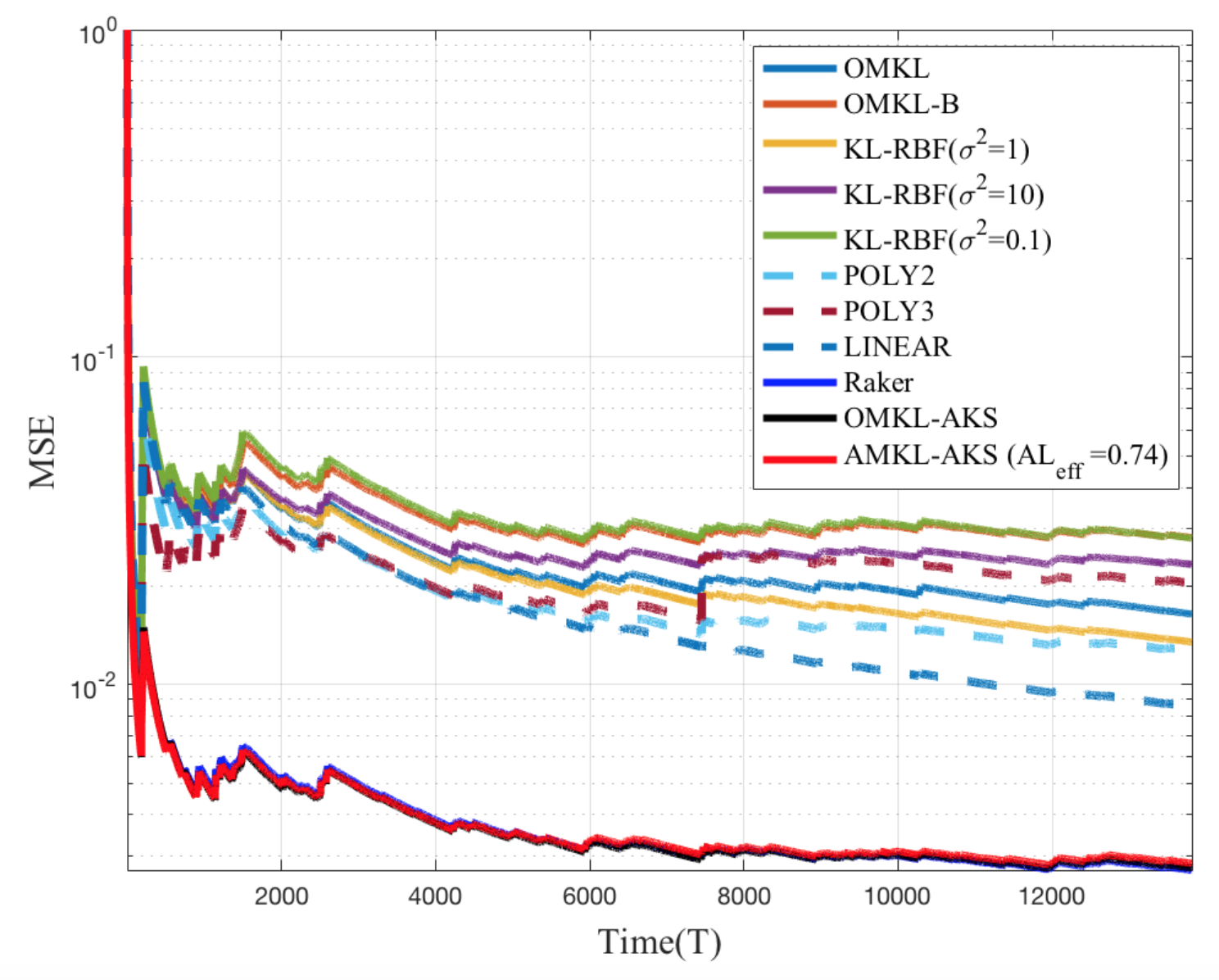}
}
\centering
\subfigure[Twitter data (Large)]{
\includegraphics[width=0.45\linewidth]{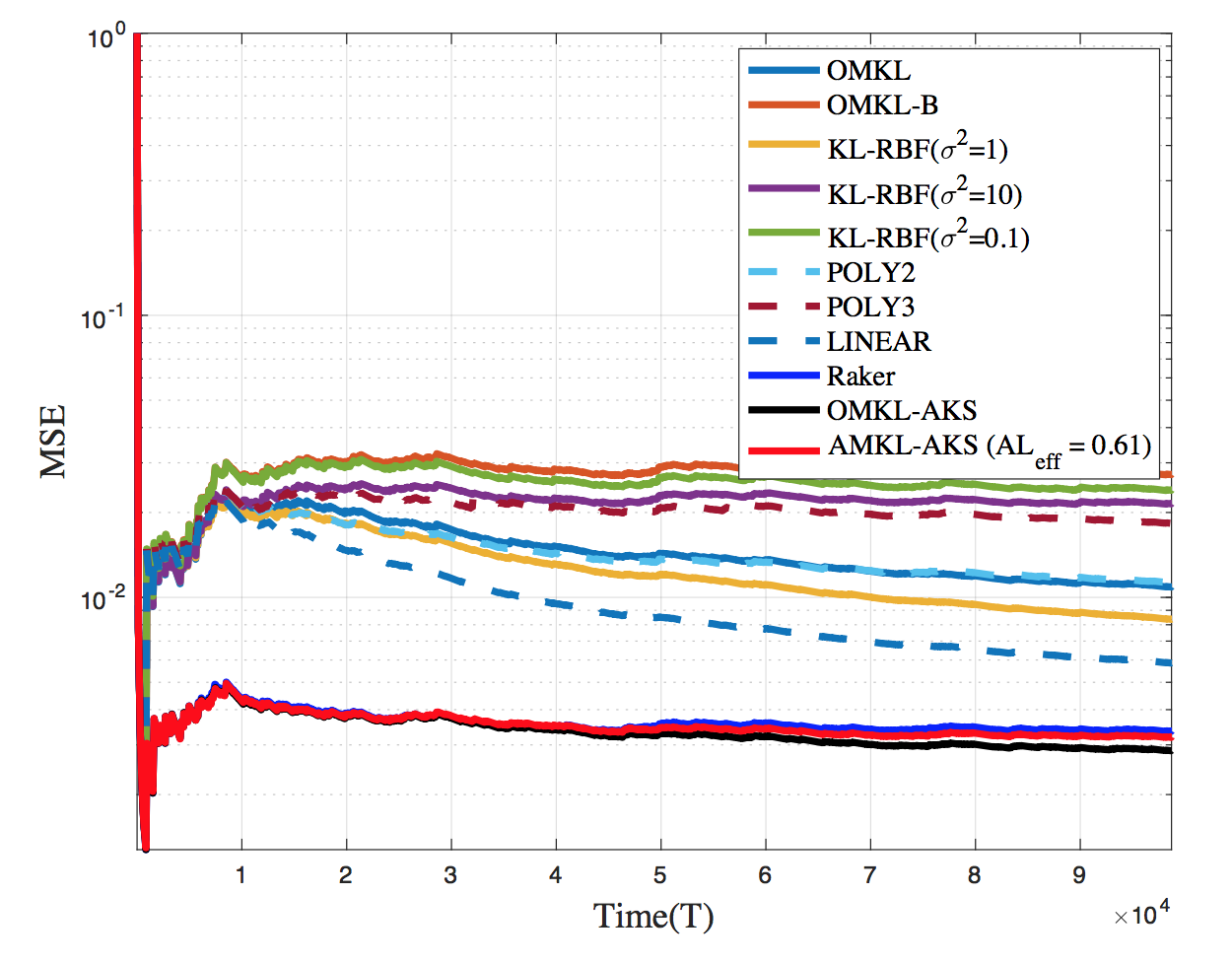}
}
\centering
\subfigure[Tom's hardware data]{
\includegraphics[width=0.45\linewidth]{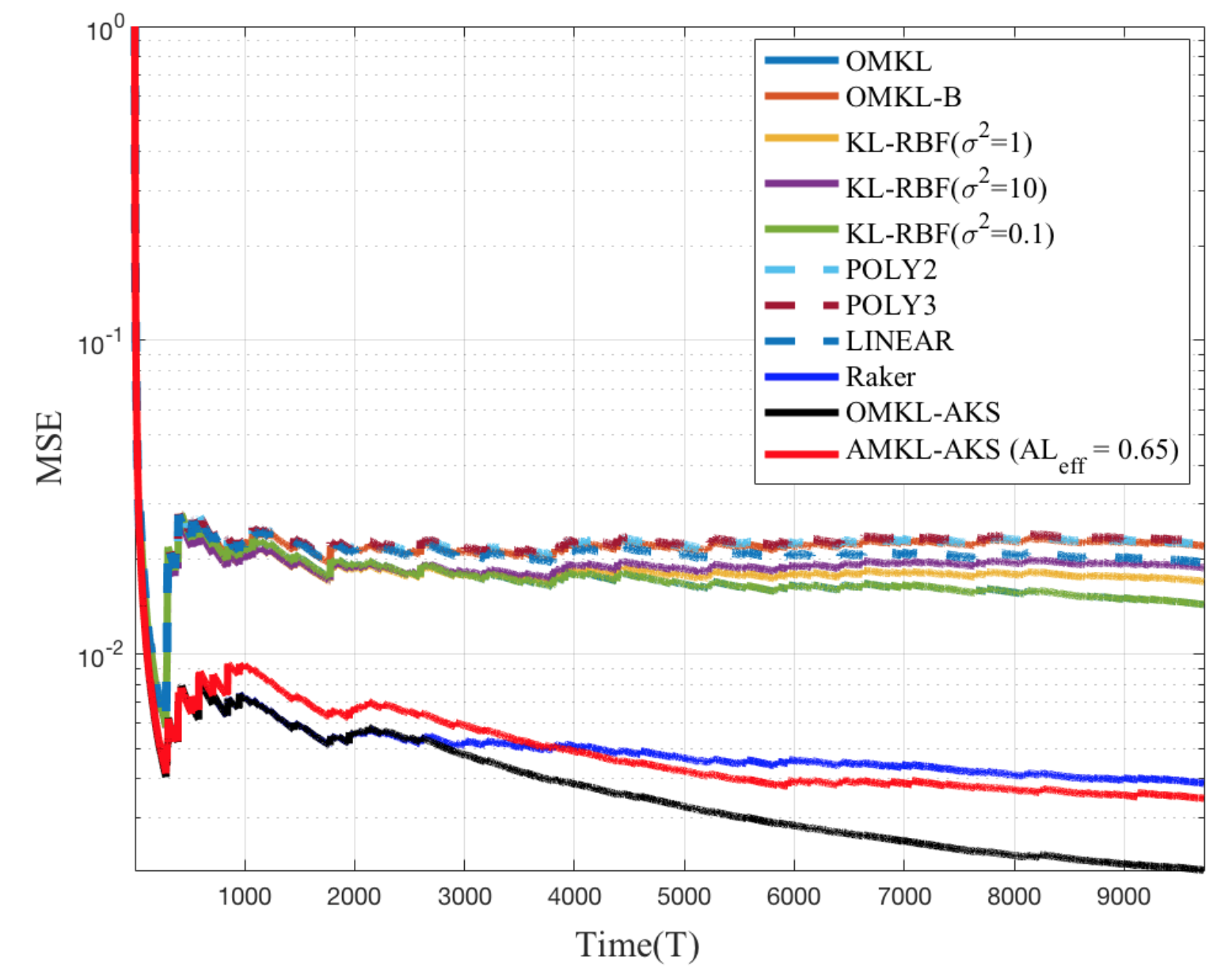}
}
\centering
\subfigure[Air quality data]{
\includegraphics[width=0.45\linewidth]{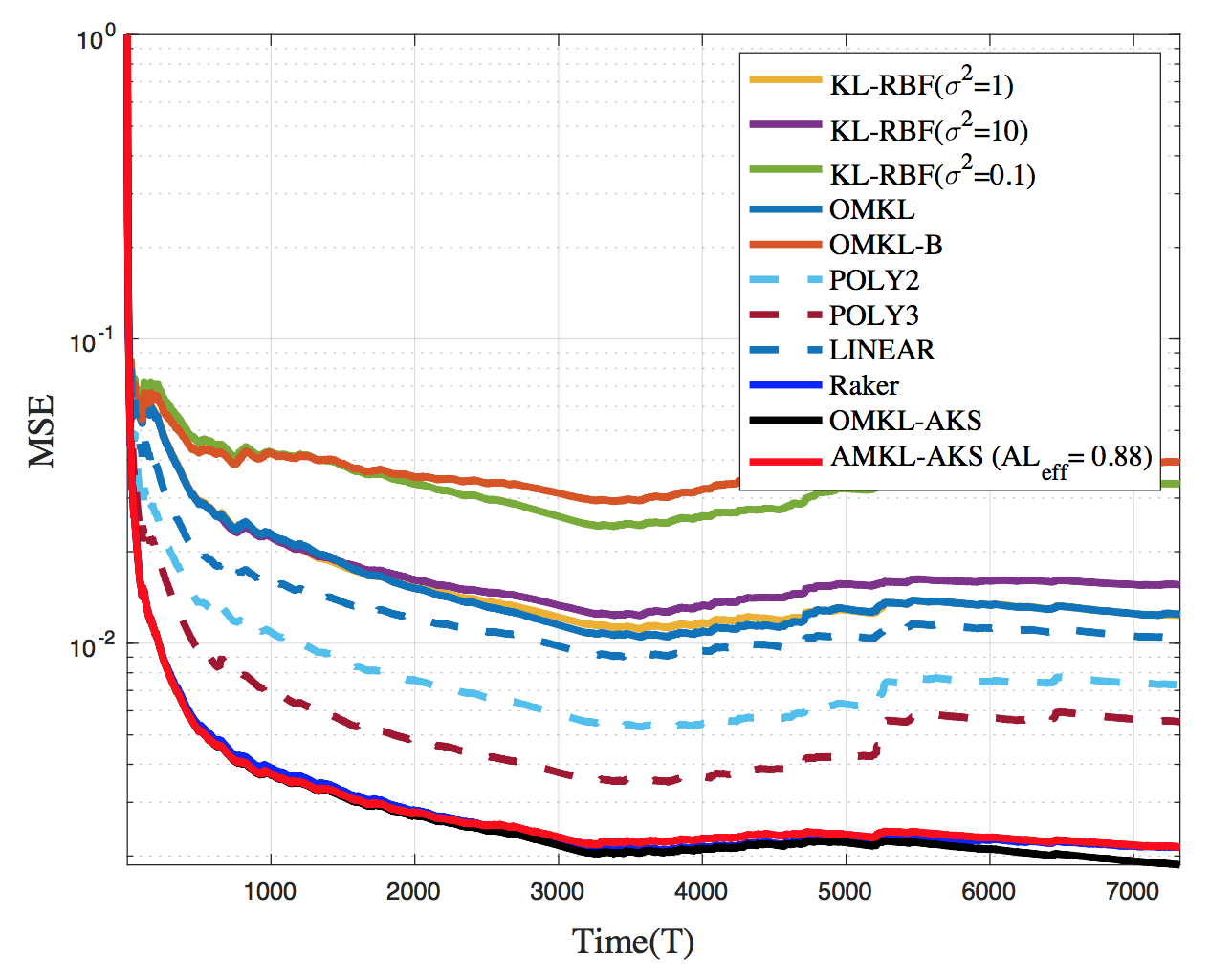}
}
\centering
\subfigure[Appliances energy data]{
\includegraphics[width=0.45\linewidth]{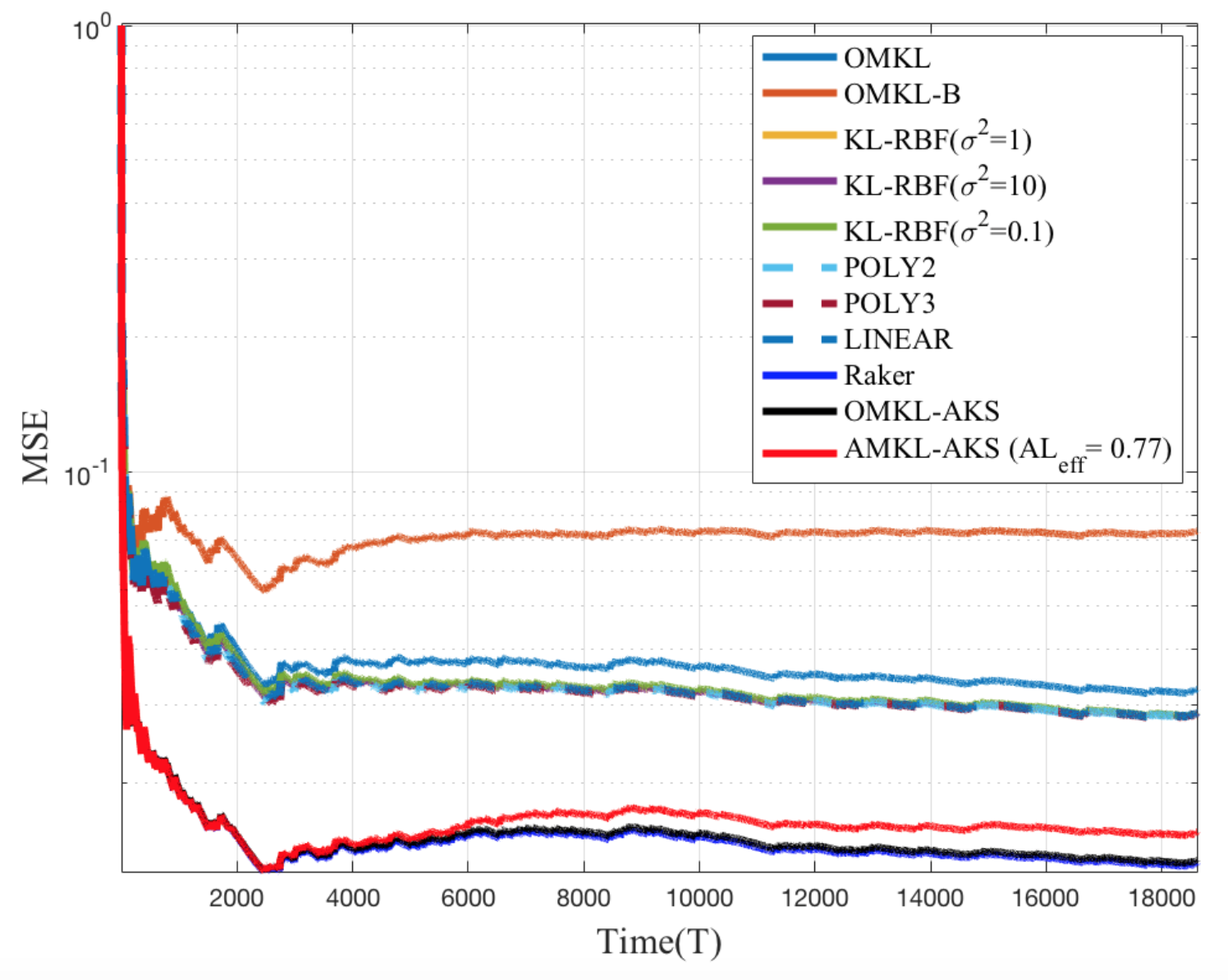}
}
\centering
\subfigure[Naval propulsion plant data]{
\includegraphics[width=0.45\linewidth]{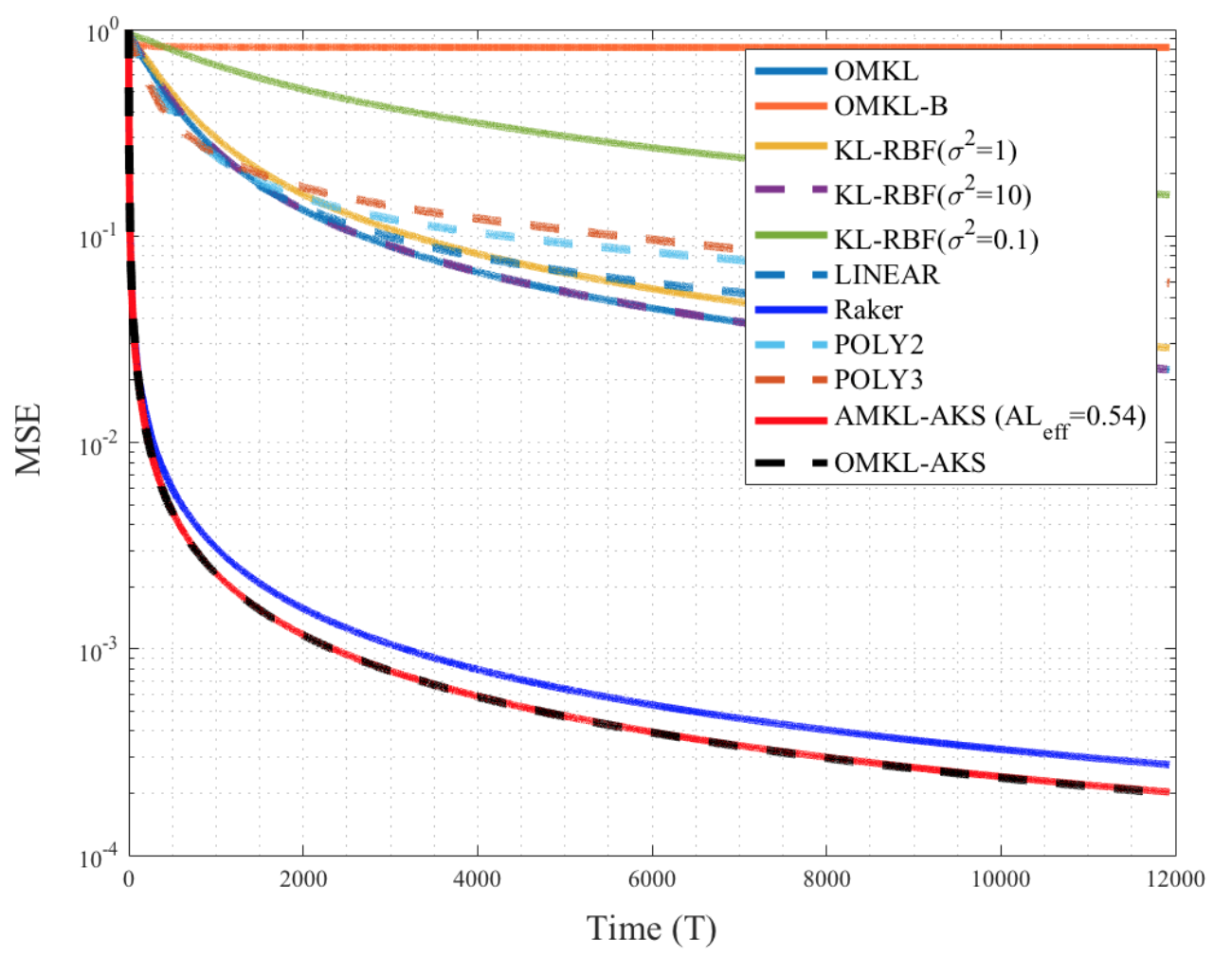}
}
\caption{MSE comparisons of various learning methods in real datasets.}
\label{fig:MSEperformance}
\end{figure*}

%
%
\section{Experiments}\label{sec:Exp}

In this section, we evaluate the performances of the proposed OMKL-AKS, AMKL, and AMKL-AKS in online regression tasks. For comparisons, we consider benchmark methods in the below:
\begin{itemize}
    \item {\bf RBF:} The online \textit{single} kernel learning method using Gaussian kernels with the parameters $\sigma^{2} = [0.1 1 10]$ (e.g., KL-RBF($\sigma^2$)).
    \item {\bf POLY:} The online \textit{single} kernel learning method using polynomial kernels with degree $d$ = \{2,3\} (e.g., POLY2 and POLY3).
    \item {\bf LINEAR:} The online \textit{single} kernel learning method using linear kernel.
    \item {\bf OMKL:} The famous online multiple kernel learning algorithm without random-feature (RF) approximation \cite{sahoo2014online}.
    \item {\bf OMKL-B:} The online multiple kernel learning algorithm on a budget \cite{kivinen2004online}.
    \item {\bf Raker:} The online multiple kernel learning algorithm based on random feature approximation \cite{shen2019random}.
\end{itemize} 
For experiments, we employ the following real datasets from UCI Machine Learning Repository. Also, they are summarized in Table~\ref{tb:DataSummary}.
\begin{itemize}
    \item {\bf Twitter \cite{Kawala2013}:} This dataset contains $T = 13818$ samples of buzz events from a famous social networks Twitter, where each attributes in ${\xv_{t} \in \RR^{77}}$ are used to predict the popularity of a topic. The larger dataset with $T$ = 98704 (termed {\bf Twitter(L)}) is included to test algorithms. 
    \item {\bf Tom's hardware \cite{Kawala2013}:} This dataset consists of $T = 9725$ samples acquired from a worldwide forum network, where each of 96 features represents such as the number of displays and the number of times a content is displayed to visitors. The task is to predict the average number of display about on a certain topic.
    \item {\bf Air quality \cite{DeVito2008}:} This dataset includes $T = 7322$ samples, which features include hourly response from an array of 5 metal oxide chemical sensors embedded in an Air Quality multi-sensor device deployed on the field in a city of Italy. The goal is to predict the concentration of polluting chemicals in the air.
    \item {\bf Appliances Energy \cite{Candanedo2017}:} This dataset contains $T = 18604$ samples describing appliances energy use such as temperature, humidity and pressure in houses. The goal is to predict energy use in a low energy building. 
    \item {\bf Naval Propulsion Plants \cite{Coraddu2014}:} This dataset has been obtained from Gas Turbine (GT) propulsion plant. Dataset contains $T = 11934$ samples with 16 features such as ship speed and fuel flow. The goal is to determine turbine decay state coefficient.
\end{itemize} 

To show the effectiveness of the proposed active learning methods, we consider the the mean-square-error (MSE) and active-learning-efficiency (${\rm AL_{eff}}$) as performance measures. They are respectively defined as
\begin{equation}
    {\rm MSE}(t)=\frac{1}{t}\sum_{\tau=1}^{t} (\hat{y}_{\tau} - y_{\tau})^2,\label{eq:mea1}
\end{equation} where $\hat{y}_{\tau}$ denotes an estimated label from the proposed algorithms and $y_{\tau}$ denotes a true label, and
\begin{equation}
    {\rm AL_{eff}}  = \frac{\text{Number of labeled samples}}{\text{Total number of samples} } = \frac{\sum_{t=1}^{T} a_t}{T}.\label{eq:mea2}
\end{equation}

Regarding the proposed algorithms and benchmark methods in the above, the following parameters will be used throughout experiments. Our parameter settings closely follow the most related paper in \cite{shen2019random} for fair comparisons. For all MKL algorithms as OMKL, OMKL-B, Raker, OMKL-AKS, AMKL, and AMKL-AKS, we use the kernel dictionary consisting of 17 Gaussian kernels whose parameters are given as
\begin{equation}
    \sigma_i^2 = 10^{\frac{i-9}{2}},\;\; i=1,...,17.
\end{equation} Also, for RF-based OMKL algorithms such as Raker, OMKL-AKS, AMKL, and AMKL-AKS, the associated parameters are set by
\begin{equation}
    \eta_l = \eta_g = \frac{1}{\sqrt{T}}, D=50, \mbox{ and } \lambda = 0.01.
\end{equation} The budget size of OMKL-K is determined as $B=50$. In OMKL-AKS and AMKL-AKS, the size-$K_t$ subset from the kernel dictionary is selected at every time $t$, where $K_t$ is determined from (\ref{eq:d_r_t}) and (\ref{eq:d_r_t_1}) for OMKL-AKS and AMKL-AKS, respectively, with $\delta_t = 0.8$ for all $t\in[T]$ and 
\begin{equation}
    \gamma_t = \min\left\{{P\choose K_t}/P, 2\right\}.
\end{equation} In this way, the size of a collection is manageable during experiments as it is always less than or equal to $\gamma_t P=34$. Finally for AMKL and AMKL-AKS, we choose the following parameters for the proposed selection criterion:
\begin{equation}
\eta_c = 0.0005 \mbox{ and } M=1.
\end{equation} Obviously, these parameters can control the tradeoff between accuracy and efficiency of AMKL-AKS, as shown in Fig.~\ref{fig:Eff-Acc}. Instead of optimizing them for each dataset, one pair of the parameters are only used for all test datasets because such kind of optimization is not practical.

\begin{table*}
\caption{Comparisons of MSE ($\times 10^{-3}$) performances.}
\label{tb:MSEperformance}
\begin{center}
\begin{tabular}{ c||c|c|c|c|c|c } 
\hline
Algorithms / Data sets & Twitter & Twitter(L) & Tom's & Air & Energy &Plant \\
\hline
KL-RBF ($\sigma^{2}$=0.1) & 28.0 & 24.0 & 14.36 & 33.37 & 28.37 & 157.9 \\ 
\hline
KL-RBF ($\sigma^{2}$=1) & 13.5 & 8.36 & 16.82 & 11.19 & 28.29 &28.68 \\ 
\hline
KL-RBF ($\sigma^{2}$=10) & 23.28 & 21.48 & 18.82 & 15.57 & 28.29 &22.52 \\ 
\hline
POLY2 & 12.68 & 11.3  & 22.73 & 7.28 & 28.21 &57.05 \\ 
\hline
POLY3 & 20.45 & 18.3  & 22.67 & 5.52 & 28.17 &59.28 \\ 
\hline
LINEAR & 8.57 & 5.85  & 19.52 & 10.43 & 28.25 &38.48 \\ 
\hline
OMKL & 16.44 & 10.9 & 13.84 & 12.46 & 31.87 & 22.49\\
\hline
OMKL-B ($B$=50) & 27.0 & 22.28 & 39.4 & 29.28 & 73.82 &818.74  \\ 
\hline
RaKer ($D$=50) & 2.72 & 3.35 & 3.87 & 2.13 & 13.4 & 0.25 \\ 
\hline
{\bf OMKL-AKS} & 2.72 & 2.87 & 2.02 & 1.87 &13.4 & 0.19 \\ 
\hline
{\bf AMKL-AKS} & \textcolor{red}{2.85}   &\textcolor{red}{3.18} & \textcolor{red}{3.46} & \textcolor{red}{2.14} & \textcolor{red}{15.3} & \textcolor{red}{0.20} \\ 
\hline
Efficiency (${\rm AL_{eff}}$) & \textcolor{blue}{0.74} & \textcolor{blue}{0.61}  & \textcolor{blue}{0.65} & \textcolor{blue}{0.88} & \textcolor{blue}{0.77} & \textcolor{blue}{0.54} \\
\hline
\end{tabular}
\end{center}
\end{table*}

\begin{table*}[t]
\caption{Comparisons of MSE ($\times 10^{-3}$) and efficiency for RF-based OMKL methods.}
\label{tb:Eff-AKS}
\begin{center}
\begin{tabular}{l|c|c|c|c|c|c|c|c|c|c|c|c}
\hline
         & \multicolumn{2}{c|}{Twitter}& \multicolumn{2}{c|}{Twitter(L)}& \multicolumn{2}{c|}{Tom's}& \multicolumn{2}{c|}{Air} &\multicolumn{2}{c|}{Energy}& \multicolumn{2}{c}{Plant}\\ \hline
         & \multicolumn{1}{l|}{MSE} & \multicolumn{1}{l|}{${\rm AL_{eff}}$} & \multicolumn{1}{l|}{MSE} & \multicolumn{1}{l|}{${\rm AL_{eff}}$} & \multicolumn{1}{l|}{MSE} & \multicolumn{1}{l|}{${\rm AL_{eff}}$} & \multicolumn{1}{l|}{MSE} & \multicolumn{1}{l|}{${\rm AL_{eff}}$} & \multicolumn{1}{l|}{MSE} & \multicolumn{1}{l|}{${\rm AL_{eff}}$} & \multicolumn{1}{l|}{MSE} & \multicolumn{1}{l}{${\rm AL_{eff}}$} \\ \hline
Raker    & 2.72 & 1 & 3.35 & 1 & 3.87 & 1 & 2.13& 1 & 13.4 & 1& 0.25& 1 \\ \hline
AMKL     & 2.67& \textcolor{blue}{0.99}& 3.32& \textcolor{blue}{0.97} & 4.26& \textcolor{blue}{0.88} & 2.13 & \textcolor{blue}{1} & 13.17 & \textcolor{blue}{1} & 0.27 & \textcolor{blue}{1} \\ \hline
OMKL-AKS & 2.72 & 1 & 2.87 & 1& 2.02 & 1& 1.87 & 1 & 13.4 & 1& 0.19& 1\\ \hline
AMKL-AKS & 2.85& \textcolor{red}{0.74} & 3.18& \textcolor{red}{0.61} & 3.46 & \textcolor{red}{0.65} & 2.14 & \textcolor{red}{0.88} & 15.3 & \textcolor{red}{0.77} & 0.20 & \textcolor{red}{0.54}\\ \hline
\end{tabular}
\end{center}
\end{table*}

\vspace{0.1cm}
\noindent {\bf Performance evaluation:} We compare the performances of the proposed OMKL-AKS, AMKL, AMKL-AKS, and the benchmark methods for real datasets in Table ~\ref{tb:DataSummary}. As performance measures,  the accuracy of a function learning in (\ref{eq:mea1}) and the efficiency of active-learning in (\ref{eq:mea2}) are considered.

Fig.~\ref{fig:MSEperformance} shows the MSE performances of various online and active learning algorithms. Also, they are summarized in Table~\ref{tb:MSEperformance} where each entry represents the MSE at the end of time. First of all, we observe that RF-based OMKL methods as Raker, OMKL-AKS, AMKL, and AMKL-AKS, significantly outperform the famous (O)MKL methods as well as single-kernel methods (e.g., Gaussian, POLY, and Linear methods). As investigated in  \cite{shen2019random}, they have much lower complexity than the other methods especially when the number of data samples is large. Since the complexity comparisons have been extensively studied in \cite{shen2019random}, such comparisons are not duplicated in this paper. Instead, we highlight the elegant accuracy-efficiency tradeoff of the proposed active learning methods. In (stream-based) active learning frameworks, the following two factors play a crucial role in determining a performance: one is to set a sharp selection criterion which enables an algorithm to select essential data samples for labeling, and the other is to exploit most relevant kernels for predicting a label precisely. In Table~\ref{tb:MSEperformance}, AMKL-AKS provides the remarkable performances, showing its solidity in terms of the above two factors. We remark that AMKL-AKS shows the comparable MSE performances with OMKL-AKS for all test datasets, which can ensure that the proposed selection criterion is accurate enough to identify unnecessary data samples. Also, from Table~\ref{tb:Eff-AKS}, we observe that AMKL-AKS performs better than AMKL (without using a kernel selection) having a smaller number of data samples in most of test datasets. These results imply that the elimination of irrelevant kernels `on the fly' (i.e., in a data-driven way) has a positive impact on AMKL, thereby enhancing the accuracy of a function learning task. In comparison of Raker and OMKL-AKS, the similar impact is observed.

More focusing on active-learning efficiency, Table~\ref{tb:MSEperformance} clearly shows that the goal of AMKL-AKS is attained as it yields highly close performances with OMKL-AKS, with $0.6\sim0.7$ efficiency on average. Remarkable, AMKL-AKS achieves the almost same performance with OMKL-AKS, by leveraging $50\%$ of labeled data. From Fig.~\ref{fig:MSEperformance}, we observe that  AMKL-AKS significantly outperforms the other methods 
(e.g. (O)MKL and POLY, KL-RBF) with a smaller number of labeled data. Based on these results, we emphasize that the proposed AMKL-AKS can have a significant impact on economical aspect having about  $30\%\sim40\%$ cost reduction for acquiring true labels. 

\begin{remark} We shed light on an important relationship between an adaptive kernel selection and active-learning efficiency. Throughout the experiment results, it is observed that the proposed kernel selection has its crucial role in enhancing the active-learning efficiency. Table~\ref{tb:Eff-AKS} shows that AMKL-AKS attains a similar or better MSE performances than AMKL (using the entire 17 kernels) with higher efficiency in all test datasets. This interesting observation leads us to conclude that AMKL-AKS indeed enjoys the advantage of refined kernels. Specifically, the kernel selection enables to improve the accuracy of the proposed selection criterion (for active labeling) as inaccurate information from irrelevant kernels can be excluded at every time. In other words, AMKL-AKS can predict a function with higher accuracy by removing irrelevant kernels, so that it is in need of just few samples compared with AMKL. \hfill$\Diamond$
\end{remark}

%
%
\section{Concluding Remarks}\label{sec:con}

In this paper, we proposed a stream-based (or sequential) active learning for online multiple kernel learning (OMKL) frameworks. The proposed method is named {\em active multiple kernel learning} (AMKL). We further improved the accuracy and efficiency of AMKL by presenting an adaptive kernel selection, which is called AMKL-AKS. Theoretically, we proved that AMKL-AKS achieves an optimal sublinear regret, implying that the proposed selection criterion indeed avoids unnecessary label-requests. Beyond asymptotic analysis, numerical tests with real datasets verified that AMKL-AKS attains a similar or better accuracy than the best-known method (termed Raker) using a smaller number of labeled data. Therefore, the proposed AMKL-AKS can yield an elegant accuracy-efficiency tradeoff. One interesting extension is to enhance AMKL-AKS by exploiting a priori knowledge on a kernel dictionary. For example, in addition to the accumulated loss information, kernel-dependencies can be also used for an adaptive kernel selection. Another extension is to develop an active learning for online graph learning frameworks, in which the graph dependencies of data samples can be used for an active labeling.



%

\appendices
\section{Proof of Lemma~\ref{lem2}}\label{app:lem2}
Recall that the weights $\hat{p}_t(i)$'s are defined in (\ref{eq:exp_st}). We let
\begin{equation*}
    \zeta = \sum_{t=1}^{T}\log\left(\sum_{i=1}^{P}\hat{p}_t(i)\exp\left(-\eta_g \Lc(\hat{f}_{t,i} (\xv_{t}),y_{t})\right)\right).
\end{equation*} The proof will be complete using the upper and lower bounds of $\zeta$. We first derive the upper bound on $\zeta$ as follows:
\begin{align}
    \zeta & \stackrel{(a)}{=}\sum_{t=1}^{T}\log\left(\EE_{I}[\exp(-\eta_g \Lc(\hat{f}_{t,I} (\xv_{t}),y_{t}))]\right)\nonumber\\
    & \stackrel{(b)}{\leq} \sum_{t=1}^{T} -\eta_g~\EE_{I}\left[\Lc(\hat{f}_{t,I} (\xv_{t}),y_{t})\right] + \frac{\eta_{g}^{2}T\ell_{u}^2}{8},\label{eq:lem2_upper}
\end{align} where (a) is due to the fact that $\pv_t$ is the PMF and (b) follows the Hoeffding's inequality with the bounded random variable $\Lc(\hat{f}_{t,I} (\xv_{t}),y_{t})$. Next, we derive the lower bound on $\zeta$ as follows:
\begin{align}
    \zeta &= \sum_{t=1}^{T}\log\left(\sum_{i=1}^{P}\hat{p}_t(i)\exp\left(-\eta_g \Lc(\hat{f}_{t,i} (\xv_{t}),y_{t})\right)\right)\nonumber\\
    &\stackrel{(a)}{=}\sum_{t=1}^T \log\left(\frac{\sum_{i=1}^{P}\hat{w}_{t+1}(i)}{\sum_{i=1}^{P}\hat{w}_t(i)}\right)\nonumber\\
    &\stackrel{(b)}{=}\log\left(\sum_{i=1}^{P} \hat{w}_{T+1}(i)\right) - \log\left(\sum_{i=1}^{P} \hat{w}_t(i)\right)\nonumber\\
    &\geq \log\left(\max_{1\leq i \leq P} \hat{w}_{T+1}(i)\right) - \log{P}\nonumber\\
    &= -\eta_{g} \min_{1 \leq i \leq P} \sum_{t=1}^{T} \Lc(\hat{f}_{t,I} (\xv_{t}),y_{t}) -\log{P}, \label{eq:lem2_lower}
\end{align} where (a) follows the definitions of $\hat{p}_t(i)$ and $\hat{w}_t(i)$ in (\ref{eq:exp_st}) and (\ref{eq:o_w}), respectively, and (b) is due to the telescoping sum. From (\ref{eq:lem2_upper}) and (\ref{eq:lem2_lower}), we can get:
\begin{align}
    &-\eta_{g} \min_{1 \leq i \leq P} \sum_{t=1}^{T}\Lc(\hat{f}_{t,I} (\xv_{t}),y_{t}) - \log{P}\nonumber\\ 
    &\;\;\;\;\;\;\;\;\;\;\;\;\;\;\; \leq \sum_{t=1}^{T} -\eta_{g}~\EE_{I}\left[\Lc(\hat{f}_{t,I} (\xv_{t}),y_{t})\right] + \frac{\eta_{g}^{2}T\ell_{u}^2}{8}.\label{eq:lem-temp10}
\end{align}The proof is completed by rearranging (\ref{eq:lem-temp10}) and  using the fact that $I\sim (\hat{p}_{t}(1),...,\hat{p}_{t}(P))$.

\section{Proof of Lemma~\ref{lem3}}\label{app:lem3}
Let $\Ac$ be the index set of revealed labels. Then, the regret can be decomposed as
\begin{align*}
    &\underbrace{\sum_{t\in\Ac}\mathcal{L}\left(\bar{f}_{t,i} ({\bf x}_{t}),y_{t}\right)- \sum_{t\in\Ac} \mathcal{L}\left(f_{i}^{\star}({\bf x}_{t}),y_{t}\right)}_{\eqdef (\star)} \\
    &\;\;\;\;\;\;\;\;\;\;\;\;\;\;\;\;\;\;\; +  \underbrace{\sum_{t\in\Ac^c}\mathcal{L}\left(\bar{f}_{t,i} ({\bf x}_{t}),y_{t}\right)- \sum_{t\in\Ac^c} \mathcal{L}\left(f_{i}^{\star}({\bf x}_{t}),y_{t}\right)}_{\eqdef (\star\star)}.
\end{align*} Clearly, the part $(\star)$ is the regret of the usual online gradient descent with time indices belong to $\Ac$. Then, from Lemma~\ref{lem1}, we can get:
\begin{equation}
    (\star) \leq \frac{C^2}{2\eta_1} + \frac{\eta_l L^2 |\Ac|}{2}.  \label{eq:lem3-part1}
\end{equation} Now we focus on the part $(\star\star)$, which is different from the usual online gradient descent. Let $\Ac^c\eqdef\{t_1,...,t_{|\Ac^c|}\}$ with $t_1<t_2<\cdots<t_{|\Ac^c|}$. Let $\Ac_{n}^c$ be the subset of $\Ac^c$ only containing non-consecutive indices, where among consecutive indices, the maximum index ins only included.  For example, if $\Ac^c=\{3,4,5,9,11,12,15\}$, then we have $\Ac_{n}^c=\{5,9,12,15\}$. Following this notation, we let
\begin{equation}
    \Ac_{n}^c=\{t_{\ell_1},...,t_{\ell_{|\Ac_{n}^c|}}\},
\end{equation} with $t_{\ell_1}<t_{\ell_2}<\cdots<t_{\ell_{|\Ac_n^c}}$.  To simplify the notation, we let $\nabla_{t}\eqdef\nabla\Lc(\bar{\thetav}_{i,t}^{\trasp}\zv_i(\xv_{t}),y_t)$. For any $t_{\ell_j}, t_{\ell_{j+1}}  \in \Ac_{n}^c$, we define the index set as
\begin{equation}
    \Tc_j=\{t_{\ell_j}+1,...,t_{\ell_{j+1}}-1\}\cap\Ac.
\end{equation} Using this, we have the following bound:
\begin{align}
    &\|\bar{\thetav}_{i,t_{\ell_{j+1}}}  - \thetav_{i}^{\star} \|^2 \nonumber\\
    &= \left\|\bar{\thetav}_{i,t_{\ell_j}} - \eta_{l}\nabla_{t_{\ell_j}} - \eta_{l}\sum_{t\in \Tc_j}\nabla_t-\thetav_{i}^{\star} \right\|^2\label{eq:temp1}\\
    &\leq \left\|\bar{\thetav}_{i,t_{\ell_{j}}} - \eta_{l} \nabla_{t_{\ell_j}} - \thetav^{\star} \right\|^2 + \eta_{l}^2\sum_{t\in \Tc_j}\|\nabla_t\|^2\nonumber\\
    &\stackrel{(a)}{\leq}  \left\|\bar{\thetav}_{i,t_{\ell_{j}}} - \eta_{l}  \nabla_{t_{\ell_j}} - \thetav_{i}^{\star}\right\|^2 + |\Tc_j|\eta_{l}^2L^2 \nonumber\\
    &= \|\thetav_{i,t_{\ell_{j}}} - \thetav_{i}^{\star}\|^2 + \eta_{l}^2\|\nabla_{t_{\ell_j}} \|^2  \nonumber\\
    &\;\;\;\;\;\;\;\;\;\;\;\;\;\;\; - 2\eta_{l} \nabla_{t_{\ell_j}}^{\trasp}(\thetav_{i,t_{\ell_j}}-\thetav_{i}^{\star}) + |\Tc_j|\eta_{l}^2L^2, \label{eq:lem3-1}
\end{align} where (a) is due to the fact that loss functions are $L$-Lipschitz.
Note that the above inequality always holds since it is  ensured that $t_{\ell_j}$ and $t_{\ell_{j+1}}$ are not consecutive. By rearranging (\ref{eq:lem3-1}), we obtain the following bound:
\begin{align}
    &\nabla_{t_{\ell_j}}^{\trasp}(\bar{\thetav}_{i,t_{\ell_j}}-\thetav_{i}^{\star})\nonumber\\
    &\leq \frac{\|\bar{\thetav}_{i,t_{\ell_j}} - \thetav_{i}^{\star}\|^2 - \|\bar{\thetav}_{i,t_{\ell_{j+1}}}  - \thetav_{i}^{\star} \|^2}{2\eta_{l}} + \frac{\eta_{l}\| \nabla_{t_{\ell_j}}^{\trasp} \|^2}{2}\nonumber\\
    &+|\Tc_j|\eta_{l}^2L^2. \label{eq:bound1}
\end{align} Also, the convexity of the loss function implies that
\begin{align*}
    &\Lc(\bar{\thetav}_{i,t_{\ell_j}}^{\trasp}\zv_i(\xv_{t_{\ell_j}}), y_{t_{\ell_j}}) - \Lc((\thetav_{i}^{\star})^{\trasp}\zv_{i}(\xv_{t_{\ell_j}}), y_{t_{\ell_j}})\\
    &\;\;\;\;\; \leq \nabla_{t_{\ell_j}}^{\trasp}(\bar{\thetav}_{i,t_{\ell_j}}-\thetav_{i}^{\star})\\
    &\;\;\;\;\; \leq \frac{\|\bar{\thetav}_{i,t_{\ell_j}} - \thetav_{i}^{\star}\|^2 - \|\bar{\thetav}_{i,t_{\ell_{j+1}}}  - \thetav_{i}^{\star} \|^2}{2\eta_{l}} + \frac{\eta_{l}\| \nabla_{t_{\ell_j}}^{\trasp} \|^2}{2} \\
    &\;\;\;\;\; +|\Tc_j|\eta_{l}^2L^2,
\end{align*} where the second inequality follows from (\ref{eq:bound1}). By telescoping sum over $t\in\Ac_{n}^c$, we can get:
\begin{align}
    &\sum_{t \in \Ac_{n}^c} \Lc(\thetav_{i,t}^{\trasp}(\zv_{i}(\xv_t)) - \sum_{t\in\Ac_{n}^c}\Lc((\thetav_{i}^{\star})^{\trasp}\zv_{i}(\xv_t))\nonumber\\
    &\;\;\;\;\; \leq \frac{\|\thetav_{i,t_{\ell_1}}-\thetav^{\star}_{i}\|^2 - \|\thetav_{i,t_{\ell_|\Ac_n^c|}}-\thetav^{\star}_{i}\|^2}{2\eta_1} + \frac{\eta_1L^2|\Ac_n^c|}{2} \nonumber\\
    &\;\;\;\;\; + \frac{\eta_1L^2}{2}\sum_{j=1}^{|\Ac_n^c|} |\Tc_j|\leq \frac{C^2}{2\eta_1}+ \frac{\eta_{l}^2L^2 T}{2},\label{eq:lem3-part2}
\end{align}where the last-inequality is due to the fact that $|\Ac_n^c|+\sum_{j=1}^{|\Ac_n^c|} |\Tc_{j}| \leq T$. By following the above procedures with the indices 
belong to $\Ac^c\setminus\Ac^{c}_n$, we have a similar bound. Since the number of consecutive unlabeling data is less than or equal to $M$, the following bound holds:
\begin{align}
    &\sum_{t \in \Ac^c} \Lc(\thetav_{i,t}^{\trasp}(\zv_{i}(\xv_t)) - \sum_{t\in\Ac^c}\Lc((\thetav_{i}^{\star})^{\trasp}\zv_{i}(\xv_t))\nonumber\\
    &\;\; \leq M\left(\frac{C^2}{2\eta_1} +  \frac{\eta_{l}^2L^2 T}{2}\right).\label{eq:lem3-part3}
\end{align} From (\ref{eq:lem3-part1}) and (\ref{eq:lem3-part3}), the proof is completed.

\section{Proofs of Lemma~\ref{lem4} and Lemma~\ref{lem5}}\label{app:lem4}

We first prove Lemma~\ref{lem4}. Leveraging the convexity of the loss function, we have:
\begin{align*}
    &\Lc(\bar{f}_t(\xv_t),\tilde{f}_t(\xv_t)) \nonumber\\
    &= \Lc\left(\sum_{i=1}^{P}\bar{p}_{t}(i) \bar{f}_{t,i} (\xv_t), \tilde{f}_t(\xv_t) \right)\\
    &\leq \sum_{i=1}^{P}\bar{p}_{t}(i)\Lc(\bar{f}_{t,i} (\xv_t), \tilde{f}_t(\xv_t))\\
    &\leq \sum_{j=1}^{P}\tilde{p}_t(j)\left(\sum_{i=1}^{P}\bar{p}_{t}(i)\Lc(\bar{f}_{t,i}(\xv_t),\bar{f}_{t,j}(\xv_t))\right)\\
    &\leq \eta_{c},
\end{align*}where the last inequality follows from the confidence condition in (\ref{eq:confidence_a}). This completes the proof.

We next prove Lemma~\ref{lem5}. Since $a_t=0$, we have $\bar{p}_{t+1}(i)=\bar{p}_{t}(i)$ and $\bar{f}_{t+1,i}(\xv)=\bar{f}_{t,i}(\xv)$ for all $i\in[P]$. From them, we have:
\begin{align*}
    &\Lc(\bar{f}_{t+1}(\xv_{t+1}),\tilde{f}_{t+1}(\xv_{t+1}))\\
    &= \Lc\left(\sum_{i=1}^{P}\bar{p}_{t}(i) \bar{f}_{t,i} (\xv_{t+1}), \tilde{f}_{t+1}(\xv_{t+1}) \right)\\
    &\leq \sum_{j=1}^{P}\tilde{p}_{t+1}(j)\left(\sum_{i=1}^{P}\bar{p}_{t}(i)\Lc(\bar{f}_{t,i}(\xv_{t+1}),\bar{f}_{t,j}(\xv_{t+1}))\right)\\
    &\leq \eta_{c}B,
\end{align*}where the last inequality is due to $a_t=0$ and the assumption (a4).

\section{Proofs of Lemma~\ref{lem6}}\label{app:lem6}

Let $\Ac=\{t\in[T]: a_t = 1\}$ be the index set of the revealed labels. Then, we have
\begin{align}
{\rm regret}_{T}^{\rm a} &= \sum_{t=1}^{T}\Lc(\bar{f}_{t}(\xv_{t}),  y_t) -  \Lc(\tilde{f}_{t}(\xv_{t}) ,y_t)\nonumber\\
&\stackrel{(a)}{\leq} \sum_{t=1}^{T} \Lc(\bar{f}_{t}(\xv_t),\tilde{f}_{t}(\xv_t))\nonumber\\
&= \sum_{t\in\Ac} \Lc(\bar{f}_{t}(\xv_t),\tilde{f}_{t}(\xv_t)) + \sum_{t\in\Ac^c} \Lc(\bar{f}_{t}(\xv_t),\tilde{f}_{t}(\xv_t))\nonumber\\
&\stackrel{(b)}{\leq} \sum_{t\in\Ac} \Lc(\bar{f}_{t}(\xv_t),\tilde{f}_{t}(\xv_t)) +\eta_c |\Ac^c|, \label{eq:lem5_1}
\end{align}where (a) is due to the assumption (a5) (i.e., triangle inequality) and (b) follows from Lemma~\ref{lem4}. In the remaining part of this proof, we will show that he first term in (\ref{eq:lem5_1}) is also bounded by $\eta_c B$. Consider an arbitrary time index $t\in\Ac$ with $t_1 < t <t_2$ for $t_1,t_2 \in \Ac^c$. From Lemma~\ref{lem5}, for $t=t_1+1$, we obtain the following upper bound:
\begin{equation}
    \Lc(\bar{f}_{t_1+1}(\xv_{t_1+1}),\tilde{f}_{t_1+1}(\xv_{t_1+1}))\leq \eta_c B.
\end{equation} Also, for $t_1+1< t < t_2$ and any fixed value $\xv_t$, we have:
\begin{align}
     \Lc(\bar{f}_{t}(\xv_{t}),\tilde{f}_{t}(\xv_{t}))
     &\stackrel{(a)}{\leq} \Lc(\bar{f}_{t_1+1}(\xv_{t}),\tilde{f}_{t_1+1}(\xv_{t}))\\
     &\stackrel{(b)}{\leq} \eta_c B,
\end{align} where (a) is because the difference between $\bar{f}_{t}(\xv)$ and $\tilde{f}_{t}(\xv)$ is smaller as $ t_1+1 <t<t_2$ increases, i.e., the labeling makes them closer, and (b) is from Lemma~\ref{lem5}. From this analysis, we have:
\begin{equation}
    \sum_{t\in\Ac} \Lc(\bar{f}_{t}(\xv_t),\tilde{f}_{t}(\xv_t))\leq \eta_c B|\Ac|.\label{eq:lem5_2}
\end{equation} From  (\ref{eq:lem5_1}) and (\ref{eq:lem5_2}), we can get
\begin{equation}
    {\rm regret}_{T}^{\rm a} \leq T\eta_c B,
\end{equation} and setting $\eta_c = \mathcal{O}(1/\sqrt{T})$, the proof is completed.

\section*{Acknowledgment}
The authors would like to thank Prof. Y. Shen for encouraging feedback and sharing MKL simulations. This work was supported by the the Samsung Research Funding \& Incubation Center of Samsung Electronics (SRFC-IT1702-00).

\ifCLASSOPTIONcaptionsoff
  \newpage
\fi



%

\bibliographystyle{IEEEtran}

\begin{thebibliography}{1}


\bibitem{scholkopf2001learning}
B.scholkopf and A.J. Smola, {\em Learning with kernels: support vector machines, regularization, optimization, and beyond}. MIT press, 2001.

\bibitem{shawe2004kernel}
J. Shawe-Taylor, N. Cristianini et al., {\em Kernel methods for pattern analysis}. Cambridge university press, 2004.

\bibitem{lin2010multiple}
Y.-Y. Lin, T.-L. Liu, and C.-S. Fuh, ``Multiple kernel learning for dimensionality reduction,'' {\em IEEE Transactions on Pattern Analysis and Machine Intelligence}, vol.33, no.6, pp.1147-1160, 2010.

\bibitem{dai2016learning}
B.Dai, N.He, Y.Pan, B.Boots and L.Song, ``Learning from conditional distributions via dual embeddings,'' {\em arXiv preprint arXiv} :1607.04579, 2016.

\bibitem{rakotomamonjy2008simplemkl}
A. Rakotomamonjy, F.R.Bach, S. Canu, and Y. Grandvalet, ``Simplemkl,'' {\em Journal of Machine Learning Research}, vol.9, no. Nov, pp.2491-2521, 2008.

\bibitem{cortes2012l2}
C.Cortes, M.Mohri, and A.Rostamizadeh, ``L2 regularization for learning kernels,'' {\em arXiv preprint arXiv:} 1205.2653, 2012.

\bibitem{gonen2011multiple}
M. G{\"o}nen and E. Alpayd{\i}n, ``Multiple kernel learning algorithms,'' {\em Journal of Machine Learning Research}, vol.12, no.Jul, pp. 2211-2268, 2011.

\bibitem{bazerque2013nonparametric}
J.A. Bazerque and G.B. Giannakis, ``Nonparametric basis pursuit via sparse kernel-based learning: A unifying view with advances in blind methods,'' {\em IEEE Signal Processing Magazine}, vol.30, no. 4, pp. 112-125, 2013.

\bibitem{ma2009identifying}
J.Ma, L.K.Saul, S.Savage, and G.M.Voelker, ``Identifying suspicious urls: an application of large-scale online learning,'' in {\em Proceedings of the 26th annual international conference on machine learning}, 2009, pp.681-688.

\bibitem{richard2008online}
C.Richard, J.C.M.Bermudez, and P.Honeine, ``Online prediction of time series data with kernels,'' {\em IEEE Transactions on Signal Processing}, vol.57, no.3, pp.1058-1067, 2008.

\bibitem{kivinen2004online} 
J.Kivinen, A.J.Smola, and R.C. Williamson, ``Online learning with kernels,'' {\em IEEE Transactions on Signal Processing}, vol.52, no.8, pp.2165-2176, 2004.

\bibitem{sahoo2014online}
D.Sahoo, S.C.Hoi, and B.Li, ``Online multiple kernel regression,'' in {\em Proceedings of the 20th ACM SIGKDD international conference on Knowledge discovery and data mining}, 2014, pp.293-302.

\bibitem{shen2019random}
Y.Shen, T.Chen, and G.B. Giannakis, ``Random feature-based online multi-kernel learning in environments with unknown dynamics,'' {\em Journal of Machine Learning Research}, vol.20, no.1, pp.773-808, 2019.

\bibitem{wahba1990spline}
G.Wahba, {\em Spline models for observational data}. Siam, 1990, vol.59.

\bibitem{rahimi2008random} 
A.Rahimi and B.Recht, ``Random features for large-scale kernel machines,'' in {\em Advances in neural information processing systems}, 2008, pp.1177-1184.

\bibitem{zhu2005semi} 
X.Zhu, J.Lafferty, and R.Rosenfeld, ``Semi-supervised learning with graphs,'' Ph.D. dissertation, Carnegie Mellon University, language technologies institute, 2005

\bibitem{settles2008active}
B.Settles, M.Craven, and L.Friedland, ``Active learning with real annotation costs,'' in {\em Proceedings of the NIPS workshop on cost-sensitive learning}. Vancouver, CA, 2008, pp.1-10.

\bibitem{bordes2005fast} 
A.Bordes, S.Erekin, J.Weston, and L.Bottou, ``Fast kernel classifiers with online and active learning,'' {\em Journal of Machine Learning Research}, vol.6, Sep, pp.1579-1619, 2005.

\bibitem{sugiyama2009pool}
M.Sugiyama and S.Nakajima, ``Pool-based active learning in approximate linear regression,'' {\em Machine Learning}, vol.75,no.3,pp.249-274,2009.

\bibitem{mccallumzy1998employing}
A.K.McCallumzy and K.Nigamy, ``Employing em and pool-based active learning for text classification,'' in {\em Proc. International Conference on Machine Learning (ICML)}. Citeseer, 1998, pp.359-367.

\bibitem{smailovic2014stream}
J.Smailovi{\'c}, M.Gr{\v{c}}ar, N.Lavra{\v{c}}, and M.{\v{Z}}nidar{\v{s}}i{\v{c}}, ``Stream-based active learning for sentiment analysis in the financial domain,'' {\em Information sciences}, vol.285, pp.181-203, 2014. 

\bibitem{settles2009active}
B.Settles, ``Active learning literature survey,'' University of Wisconsin Madison Department of Computer Sciences, Tech.Rep., 2009. 

\bibitem{wu2018pool}
D.Wu, ``Pool-based sequential active learning for regression,'' {\em IEEE Transactions on neural networks and learning systems}, vol.30, no.5, pp.1348-1359, 2018.

\bibitem{dagan1995committee}
I.Dagan and S.P.Engelson, ``Committee-based sampling for training probabilistic classifiers,'' in Machine Learning Proceedings 1995. Elsevier, 1995, pp.150-157.

\bibitem{krishnamurthy2002algorithms}
V.Krishnamurthy, ``Algorithms for optimal scheduling and management of hidden markov model sensors,'' {\em IEEE Transactions on Signal Processing}, vol.50, no.6, pp. 1382-1397, 2002.

\bibitem{yu2005svm}
H.Yu, ``SVM selective sampling for ranking with application to data retrieval,'' in {\em Proceedings of the eleventh ACM SIGKDD international conference on Knowledge discovery in data mining}, 2005, pp.354-363.

\bibitem{vzliobaite2013active}
I.{\v{Z}}liobait{\.e}, A.Bifet, B.Pfahringer, and G.Holmes, ``Active learning with drifting streaming data,'' {\em IEEE Transactions on neural networks and learning systems},vol.25, no.1, pp.27-39, 2013. 

\bibitem{hao2018online}
S.Hao, P.Hu, P.Zhao, S.C.Hoi, and C.Miao, ``Online active learning with expert advice,'' ACM Transactions on Knowledge Discovery from Data (TKDD), vol.12, no.5, pp.1-22, 2018.

\bibitem{smola1998learning}
A.J. Smola and B.Scholkopf, {\em Learning with kernels}, Citeseer, 1998,vol.4.

\bibitem{micchelli2005learning}
C.A. Micchelli and M.Pontil, ``Learning the kernel fuction via regularization,'' {\em Journal of Machine Learning Research}, vol.6, no.Jul, pp.1099-1125, 2005. 

\bibitem{hazan2016introduction} 
E.Hazen {\em et al}., ``Introduction to online convex optimization,'' Foundations and Trends {\textregistered} in Optimization, vol.2, no.3-4, pp.157-325, 2016.

\bibitem{bubeck2011introduction}
S.Bubeck, ``Introduction to online optimization,'' Lecture Notres, vol.2, 2011

\bibitem{wainwright2019high}
M.J. Wainwright, High-dimensional statistics: High-dimensional statistics: A non-asymptotic viewpoint. Cambridge University Press, 2019, vol.48.

\bibitem{Kawala2013}
E.G. François Kawala, Ahlame Douzal-Chouakria and E. Dimert, ``Pr{\'e}dictions d’activit{\'e} dans les r{\'e}seaux sociaux en ligne,'' {\em 4i{\`e}me Conf{\'e}rence sur les Mod{\`e}les et l’Analyse des R{\'e}seaux: Approches Math{\'e}matiques et Informatiques}, 2013.

\bibitem{DeVito2008}
M.P.L.M. Saverio De Vito, Ettore Massera and G.D. Francia, ``On field calibration of an electronic nose for benzene estimation in an urban pollution monitoring scenario.'' {\em Sensors and Actuators B: Chemical}, vol.129, no.2, pp.750-757, 2008.

\bibitem{Candanedo2017}
V.F.Luis, M.Candanedo and D.Deramaix, ``Data driven prediction models of energy use of appliances in a low-energy house,'' {\em Energy and Buildings}, vol.140, pp.81-97, 2017.

\bibitem{Coraddu2014}
Coraddu, A., Oneto, L., Ghio, A., Savio, S., Anguita, D., and Figari, M, ``Machine learning approaches for improving condition based maintenance of naval propulsion plants,'' {\em Journal of Engineering for the Maritime Environment}, 2014.






\end{thebibliography}




%








\end{document}